\RequirePackage{fix-cm}
\documentclass[twocolumn]{svjour3}          
\smartqed  
\usepackage{natbib}
\usepackage{times}
\usepackage{epsfig}
\usepackage{graphicx}
\usepackage{subfigure}
\usepackage{amsmath}
\usepackage{amssymb}
\usepackage{color}
\usepackage{framed,multirow}
\usepackage{hhline}
\usepackage{array}
\usepackage{colortbl}
\usepackage{xcolor}
\usepackage{amssymb}
\usepackage{latexsym}
\usepackage{ragged2e}
\usepackage{makecell}
\usepackage{graphicx}
\usepackage{makecell}
\usepackage{amsmath,amssymb} 
\usepackage{color}
\usepackage{array}
\usepackage{algorithm} 
\usepackage{algorithmic} 
\usepackage{subfigure}
\usepackage{multirow}
\usepackage{slashbox}
\usepackage{mathptmx}

\usepackage[misc]{ifsym}
\usepackage{natbib}
\usepackage{pifont}
\usepackage{bbding}
\usepackage[colorlinks,
            linkcolor=blue,
            citecolor=blue
            ]{hyperref}

\begin{document}

\title{Learning Unknown Spoof Prompts for Generalized Face Anti-Spoofing Using Only Real Face Images}
\author{ Fangling Jiang \textsuperscript{\rm 1} \and
	Qi Li \textsuperscript{\rm 2,3}\and
	Weining Wang \textsuperscript{\rm 4} \and
	Wei Shen \textsuperscript{\rm 5} \and
	Bing Liu \textsuperscript{\rm 1} \and
	Zhenan Sun \textsuperscript{\rm 2,3}
}

\institute{ 
	Fangling Jiang \at
	\email{jiangfangling66@gmail.com} \and
	Qi Li {\scriptsize (\Letter)} \at
	\email{qli@nlpr.ia.ac.cn} \and
	Weining Wang \at
	\email{weining.wang@nlpr.ia.ac.cn} \and
	Wei Shen  \at
	\email{shenwei12@oppo.com} \and    
	Bing Liu \at
	\email{bingliu@usc.edu.cn} \and
		Zhenan Sun \at
	\email{znsun@nlpr.ia.ac.cn} \\
	\\
	\textsuperscript{\rm 1} School of Computer Science, University of South China, Hengyang, China\\
	\textsuperscript{\rm 2} New Laboratory of Pattern Recognition, MAIS, CASIA, Beijing, China\\
	\textsuperscript{\rm 3} School of Artificial Intelligence, UCAS, Beijing, China\\	
      \textsuperscript{\rm 4} The Laboratory of Cognition and Decision Intelligence for Complex Systems, CASIA, Beijing, China\\
       \textsuperscript{\rm 5} OPPO AI Center, Beijing, China\\
       \textsuperscript{\rm 6} Fangling Jiang and Qi Li contributed equally to this study.\\
}

\date{Received: date / Accepted: date}

\maketitle
\begin{abstract}
Face anti-spoofing is a critical technology for ensuring the security of face recognition systems. However, its ability to generalize across diverse scenarios remains a significant challenge. In this paper, we attribute the limited generalization ability to two key factors: covariate shift, which arises from external data collection variations, and semantic shift, which results from substantial differences in emerging attack types. To address both challenges, we propose a novel approach for learning unknown spoof prompts, relying solely on real face images from a single source domain. Our method generates textual prompts for real faces and potential unknown spoof attacks by leveraging the general knowledge embedded in vision-language models, thereby enhancing the model’s ability to generalize to unseen target domains. Specifically, we introduce a diverse spoof prompt optimization framework to learn effective prompts. This framework constrains unknown spoof prompts within a relaxed prior knowledge space while maximizing their distance from real face images. Moreover, it enforces semantic independence among different spoof prompts to capture a broad range of spoof patterns. Experimental results on nine datasets demonstrate that the learned prompts effectively transfer the knowledge of vision-language models, enabling state-of-the-art generalization ability against diverse unknown attack types across unseen target domains without using any spoof face images.

\end{abstract}

\section{Introduction}\label{sec1}
In recent years, face recognition technology has been extensively employed in various scenarios such as identity verification, criminal tracking, and health management. However, face recognition systems are frequently subjected to attacks involving spoof faces, such as printed photos, replayed face videos, and 3D masks. These spoof faces attempt to bypass face recognition verification and gain unauthorized access to legitimate user privileges, posing a significant threat to the security of face recognition technology. To safeguard face recognition security, face anti-spoofing (FAS), which aims to distinguish real faces from spoof ones, has garnered attention from both academia and industry.

Early face anti-spoofing methods~\citep{mmboulkenafet2015face,yi2014face} primarily rely on handcrafted descriptors such as LBP and Gabor to extract features for binary classification between real and spoof faces. Unfortunately, in practical applications, the unseen test data often exhibits covariate shift compared to the training data due to variations in extrinsic factors that are spoof-irrelevant but affect the appearance of captured images (e.g., recording devices, background settings, and lighting conditions). These approaches demonstrate poor generalization ability to such shift. To address this issue, recent research has focused on learning deep representations for face anti-spoofing, with particular emphasis on domain generalization techniques to learn domain-invariant features for unseen target domains~\citep{shao2019multi,jia2020single,cai2024towards,liu2025dual}. This advancement has significantly enhanced the generalization ability of face anti-spoofing models against covariate shift.

\begin{figure}[tp]
	\centering
	\includegraphics[width=0.5\textwidth]{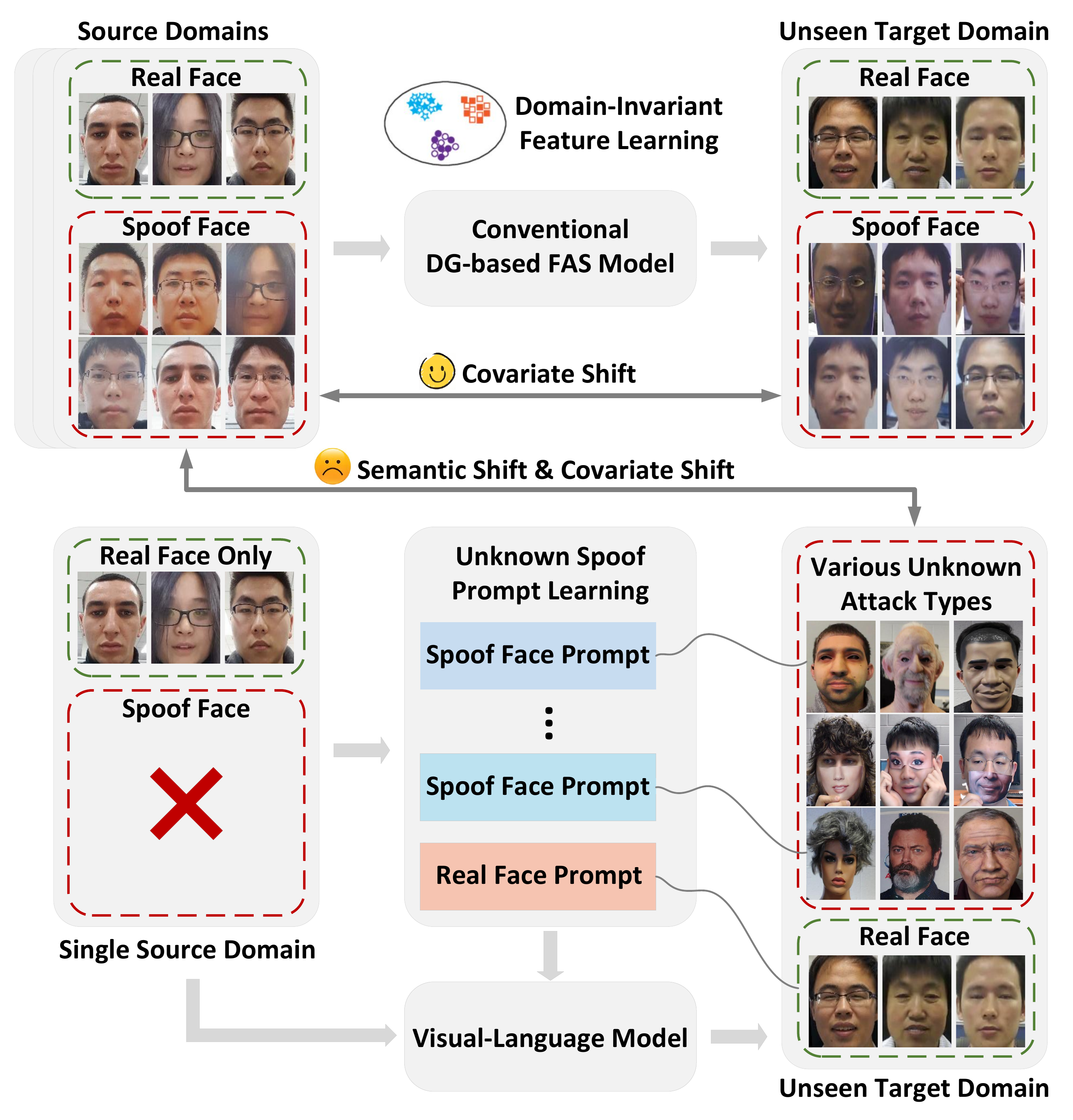}
	\caption{Conventional domain generalization (DG)-based FAS methods primarily focus on addressing covariate shift, but generalize poor to semantic shift caused by the emergence of unknown attack types. We learn potential spoof face prompts for unknown attack types solely based on real face images, adapting the pre-trained knowledge of the vision-language model to generalize well to semantic shift and covariate shift simultaneously present in unseen target domains.
	}
	\label{fig1}
\end{figure}

Existing domain generalization-based FAS methods primarily address covariate shift by leveraging multiple source domains, under the assumption that the unseen target domain shares the same attack types as the source domains. However, with advancements in manufacturing techniques and decreasing production costs, generating attack types in diverse forms has become increasingly simple and accessible~\citep{george2019biometric,guo2022multi}. Consequently, it has become increasingly challenging for the source domain to encompass all possible attack types that may appear in the unseen target domain.  This results in not only conventional covariate shift between the source and unseen target domains but also semantic shift caused by the emergence of novel spoof patterns, as illustrated in \figurename~\ref{fig1}.
The domain generalization-based FAS methods tend to overfit the attack types encountered in the source domains,  making it generalize poor to previously unseen attack types.
Since unknown attack types are inherently unpredictable, it is challenging to capture all possible variations. Meanwhile, the distributional differences of real face data across different domains are relatively small.  In this paper, we propose learning a FAS model solely based on real face images from a single source domain, without using any spoof face images, to enable low-cost yet generalized handling of both semantic and covariate shifts in unseen target domains.

Specifically, we propose a prompt learning framework that learns real prompts for real faces and infers spoof prompts for unknown attack types. This enables effective textual prompt learning, allowing the vision-language model~\citep{zhang2024vision} to adapt pre-trained knowledge for generalized classification of real faces and unknown attacks across target domains.
To achieve effective prompt learning, the following challenges need to be addressed:  
(1) how to infer potential spoof prompts solely from real face training images; 
(2) how to ensure the diversity of spoof prompts to accommodate various unknown attack types; 
(3) how to guarantee that the learned spoof prompts encapsulate the contextual information of unknown attacks rather than being arbitrarily generated; and  
(4) how to ensure the strong discriminative power of the learned prompts.

To address these challenges, we begin by focusing on real face images and employ a contrastive learning approach to extrapolate potential spoof prompts within the embedding space. We maximize the distance between spoof prompts and real face representations while constraining real prompts to remain closely with real face representations.
To ensure that the generated spoof prompts effectively capture the diversity of unknown attack types, we construct them as combinations of multiple textual prompts and further enforce semantic independence among them. Additionally, we leverage the prior knowledge embedded in large language models regarding spoof faces, ensuring that the inferred spoof prompts remain within a relaxed yet meaningful range of this prior knowledge, thereby encapsulating critical contextual information about unknown attacks. Finally, we introduce a one-class discriminative classification regularization to refine the generated prompts, minimizing the discrepancy between the predicted probability distribution of real face images and the binary ground-truth distribution. This not only explicitly enhances the discriminability of real prompts, but also implicitly reduces the similarity between spoof prompts and real face images, reinforcing the robustness of the proposed framework.

The main contributions of this paper are summarized as follows:
\begin{enumerate}
	\item We propose an innovative prompt learning framework that relies solely on real face images from a single source domain to generate potential spoof prompts for unknown attack types. By effectively adapting the pre-trained knowledge of the vision-language model, our framework achieves strong generalization across both semantic and covariate shifts in unseen target domains while maintaining low training data requirements.
	\item To improve the quality of generated spoof prompts, we propose a diverse spoof prompt optimization paradigm. This paradigm maximizes the separation between spoof prompts and real face representations under the sparse guidance of prior spoofing knowledge while ensuring semantic independence among different spoof prompts, thereby facilitating the learning of diverse and representative spoof patterns for unknown attack types.
	\item Extensive experiments on nine face anti-spoofing datasets demonstrate that our approach achieves state-of-the-art performance. Despite being trained solely on real face images from a single source domain, our method effectively defends against diverse unknown attack types in unseen target domains.
\end{enumerate}

\section{Related Work}\label{sec2}
\subsection{Domain Generalization-based Face Anti-Spoofing}
The poor cross-scenario generalization ability of models remains a significant challenge in face anti-spoofing research. To address this issue, domain generalization techniques, which aim to develop models capable of generalizing to unseen target domains by leveraging multiple related source domains~\citep{wang2022generalizing}, have emerged as the predominant approach. Existing domain generalization-based face anti-spoofing methods can be broadly categorized into five groups: domain alignment-based approaches, meta-learning-based approaches, disentangled representation learning-based approaches, prompt learning-based approaches, and other miscellaneous techniques.

\emph{a) Domain alignment-based approaches}

Domain alignment-based approaches aim to align data distributions across multiple source domains to learn domain-invariant features, thereby improving generalization to unseen target domains. These methods typically align the marginal distribution\citep{li2018learning,shao2019multi}, single-side distribution\citep{wang2024csdg,jia2020single}, or conditional distribution\citep{jiang2023adversarial} across source domains using techniques such as maximum mean discrepancy minimization\citep{li2018learning} and domain adversarial learning~\citep{shao2019multi,jia2020single,jiang2023adversarial}.
In domain adversarial learning, Kong et al.\citep{kong2024dual} argue that the competition between the feature encoder and the domain discriminator often leads to training instability and slow convergence. To mitigate this, they design a domain adversarial attack module that introduces noise into input images, reducing inter-domain differences and facilitating domain feature alignment. Additionally, traditional domain alignment methods struggle to ensure convergence toward a flat minimum. Le et al.\citep{le2024gradient} address this by performing gradient alignment at the ascending points of each source domain, ensuring convergence to the optimal flat minimum and thereby enhancing generalization to covariate shift.

Compared to conventional methods that focus on global feature alignment across source domains, Liu et al.\citep{liu2025dual} consider semantic relationships between local regions, aligning local features with rich semantic information. Sun et al.\citep{sun2023rethinking} enforce alignments of the live-to-spoof transition direction across multiple source domains, preserving key spoofing discrimination information while addressing covariate shift. Liu et al.\citep{liu2024moeit} integrate cross-attention mechanisms with MOE, achieving domain-invariant feature alignment while complementing domain-specific features for live and spoof face classification. Furthermore, Liu et al.\citep{liu2024quality} argue that face anti-spoofing models are sensitive to face quality and propose a teacher-student framework that aligns multi-source domain features within a quality-invariant space.
Since direct distribution alignment often overlooks hierarchical relationships between samples, Hu et al.~\citep{hu2024rethinking} leverage prototype learning to perform hierarchical feature alignment in hyperbolic space.

\emph{b) Meta-learning-based approaches}

Following the pioneering work of Shao et al.~\citep{shao2020regularized}, most meta-learning-based face anti-spoofing methods have adopted the leave-one-domain-out strategy on source domains to simulate covariate shift. Specifically, these methods designate one source domain as the meta-test set while using the remaining source domains as the meta-training set.
Early studies~\citep{jia2021dual} introduce deep loss and triplet loss to regularize the meta-learning optimization process, promoting the extraction of domain-shared features. However, while domain-shared feature learning enhances generalization, it may reduce feature discriminability within individual source domains.
To address this limitation, Zhou et al.\citep{zhou2022adaptive} incorporate domain-specific feature learning through a meta-learning optimization strategy that adaptively aggregates knowledge from multiple domain experts. Building on this, Muhammad et al.\citep{muhammad2023domain} propose a novel approach that first synthesizes diverse augmented images for each source domain, then trains domain-specific expert classifiers using these synthetic datasets, and finally employs a meta-learning model to integrate knowledge from multiple experts, further improving cross-domain generalization.
Beyond traditional binary discriminative models, Du et al.~\citep{du2022energy,zhang2024domain} introduce an alternative method that combines meta-learning with an energy-based model, constructing a generative framework for face anti-spoofing.

\emph{c) Disentangled representation learning-based approaches}

Disentangled representation learning assumes that learned features can be categorized into domain-shared and domain-specific features. Domain-shared features capture spoof-related, cross-domain invariant information, while domain-specific features correspond to external factors unrelated to spoofing that introduce domain variations.
Building on this concept, Wang et al.\citep{wang2022domain} propose a shuffled style assembly network that enhances domain-shared features while suppressing domain-specific features, enabling the learning of highly generalizable representations. Ma et al.\citep{ma2024dual} disrupt facial structures to decouple spoof-related features from facial attribute features. Similarly, Yang et al.~\citep{yang2024generalized} perform fine-grained domain partitioning based on data characteristics and enforce orthogonality constraints between spoof-related features and identity attribute features to achieve effective disentanglement.

\emph{d) Prompt learning-based approaches}

Vision-language models trained on large-scale datasets demonstrate strong zero-shot generalization across various downstream vision tasks~\citep{park2023visual}. Prompt learning has emerged as an effective approach for adapting the pre-trained knowledge of vision-language models to specific downstream tasks~\citep{zhang2024vision}. Common prompt learning methods include text prompt learning (e.g., CoOp~\citep{Zhou_2022}, CoCoOp~\citep{zhou2022conditional}), visual prompt learning (e.g., VPL~\citep{bahng2022exploring}), and multi-modal prompt learning~\citep{zang2022unified,khattak2023maple}. Recently, prompt learning has been introduced to advanced face anti-spoofing work, offering new possibilities for improving generalization ability.

Early attempts to integrate vision-language models into face anti-spoofing rely on fixed class prompts and full-model fine-tuning to improve accuracy and generalization~\citep{srivatsan2023flip}. However, more recent approaches have largely frozen the parameters of vision-language models, adapting them to the face anti-spoofing domain through learnable prompt learning.
A key advancement in this direction involves incorporating feature disentanglement into prompt learning. Hu et al.~\citep{hu2024fine} introduce domain-specific and domain-invariant prompts to capture both shared and specified domain information. Liu et al.~\citep{liu2024cfpl} condition prompts on content and style features, enhancing the diversity of learned semantic representations. Similarly, Mu et al.~\citep{mu2023teg} construct matched and mismatched textual prompts to enforce a cross-domain universal constraint, ensuring that the model focuses on domain-invariant features rather than domain-specific details.  

Building on these foundations, Wang et al.~\citep{wang2024tf} argue that coarse-grained prompts fail to fully exploit language as supervisory information. To address this, they propose a fine-grained prompt design that decouples content and category representations, aligning prompts more effectively with the knowledge of vision-language models. Similarly, Guo et al.\citep{guo2024style} introduce style-conditioned prompts to extract multi-level features, while Liu et al.\citep{liu2024bottom} further advance this concept by integrating domain adversarial training with fine-grained prompt learning, addressing domain discrepancies at multiple levels. In addition to structural improvements, Fang et al.~\citep{fang2024vl} refine prompt learning by incorporating fine-grained textual descriptions of facial regions, reducing the model’s focus on irrelevant facial details. 
Beyond classification, Zhang et al.~\citep{zhang2025interpretable} tackle the lack of interpretability in conventional face anti-spoofing, where models typically provide only confidence scores. They reformulate face anti-spoofing as a visual question-answering task, enabling the generation of interpretable textual explanations for real and spoof face classifications. 

\emph{e) Other miscellaneous techniques}

Existing methods primarily focus on learning generalized features from the spatial domain. In contrast, Zheng et al.~\citep{zheng2024mfae} tackle the problem from the frequency domain by randomly masking low-frequency information, which contains domain-specific details. They then employ a ViT-based encoder-decoder structure to extract domain-invariant features from the masked images, thereby enhancing model generalization.
When transferring a pre-trained ViT to the face anti-spoofing domain, the inductive bias of linear layers often degrades the model's generalization ability. To address this, Cai et al.~\citep{cai2024s} design a statistics-based adapter that extracts discriminative information from localized token histograms, mitigating the influence of linear layers and improving generalization.

Most existing work focuses on model optimization, while Cai et al.~\citep{cai2022learning,cai2024towards} explore data augmentation by generating images with spoof artifacts such as blur and moire patterns, thereby improving both the quality and quantity of training data for generalized feature learning. Similarly, Ge et al.~\citep{ge2024difffas} introduce a diffusion model to generate spoof face images corresponding to real ones, addressing the issue of insufficient training data for novel attack types.
Additionally, to handle novel attack types, some methods~\citep{long2024confidence,jiang2024cross} not only predict the probability of a testing sample belonging to a specific class but also assess the confidence of the prediction. By rejecting samples with low confidence, these approaches enhance face anti-spoofing performance in detecting unseen target domains and unknown attack types.
Despite these advancements, most domain generalization methods struggle with large covariate shift. To address this, Zhou et al.~\citep{zhou2024test} propose test-time style projection and diverse style shift simulation modules, which map unseen target domains into the known training domain space during inference, improving generalized real and spoof face classification.

In summary, most existing domain generalization-based face anti-spoofing methods rely on multiple source domains containing both real and spoof face image data for training. In contrast, the proposed method requires only real face image data from a single source domain. This significantly reduces the cost of training data, as collecting diverse spoof face image data is notoriously time-consuming, labor-intensive, and expensive. Furthermore, most existing domain generalization-based face anti-spoofing methods focus on addressing covariate shift, often struggling with semantic shift. The proposed method, however, generates fine-grained and diverse textual prompts for potential unknown attack types, effectively adapting the pre-trained knowledge of vision-language models to address both covariate and semantic shifts in unseen target domains for face anti-spoofing.

\subsection{One-Class Face Anti-Spoofing}

Existing domain generalization-based face anti-spoofing methods focus on addressing cross-domain covariate shift, but have poor generalization ability to semantic shift caused by unknown attack types. Considering the high diversity of attack types, some researchers have proposed framing face anti-spoofing as a one-class classification task. These approaches aim to improve the model’s generalization ability to unknown attack types by focusing on learning the intrinsic characteristics of real faces rather than explicitly modeling diverse spoofing attacks.

Many approaches train one-class classifiers via Gaussian mixture models~\citep{nikisins2018effectiveness}, autoencoders~\citep{abduh2020use,huang2021one}, kernel fisher null-space regression models~\citep{arashloo2020unseen} or localized multiple kernel learning models~\citep{arashloo2023unknown}
to model the data distribution of real face samples and detect unknown attacks. Moreover, Fatemifar et al.~\citep{fatemifar2020stacking,fatemifar2019combining} leverage genetic algorithms to integrate the decisions of multiple one-class classifiers, further enhancing the model's generalization ability to unknown attacks.

Since no spoof face image data is available, some methods~\citep{baweja2020anomaly,narayan2024hyp,gwon2024one} generate pseudo-spoof face samples by sampling from the Gaussian distribution of real faces for subsequent discriminative training. 
Among them, ~\citep{narayan2024hyp} further constrains the learned features in the hyperbolic space to be both discriminative for real and spoof face classification and disentangled from identity information, thereby improving the model's accuracy in detecting unknown attacks.
However, cross-domain covariate shift may cause the instances sampled based on real face statistics to fail in accurately representing the distribution of spoof faces, leading to misclassifications of test samples. 
To avoid this problem, Huang et al.~\citep{huang2024one} design randomly sampled pseudo spoof cue maps as supervision for the generator to generate zero spoof cue maps for real faces and non-zero pseudo spoof cue maps for potential spoof faces, ultimately determining whether a test sample is real or spoof based on the distribution of spoof cue map values. 

Compared to previous one-class methods, the proposed method trains the face anti-spoofing model using only real face image data too. In contrast, our approach leverages prompt learning to transfer the general knowledge of vision-language models, effectively addressing covariate shift and semantic shift. Additionally, rather than relying on random sampling, our method generates diverse textual prompts for potential spoof faces based on real face visual data and prior knowledge of large language models, incorporating contextual semantic information about unknown attack types for more effective mitigation of semantic shift.

\begin{figure*}[!t]
	\centering
	\includegraphics[width=1\textwidth]{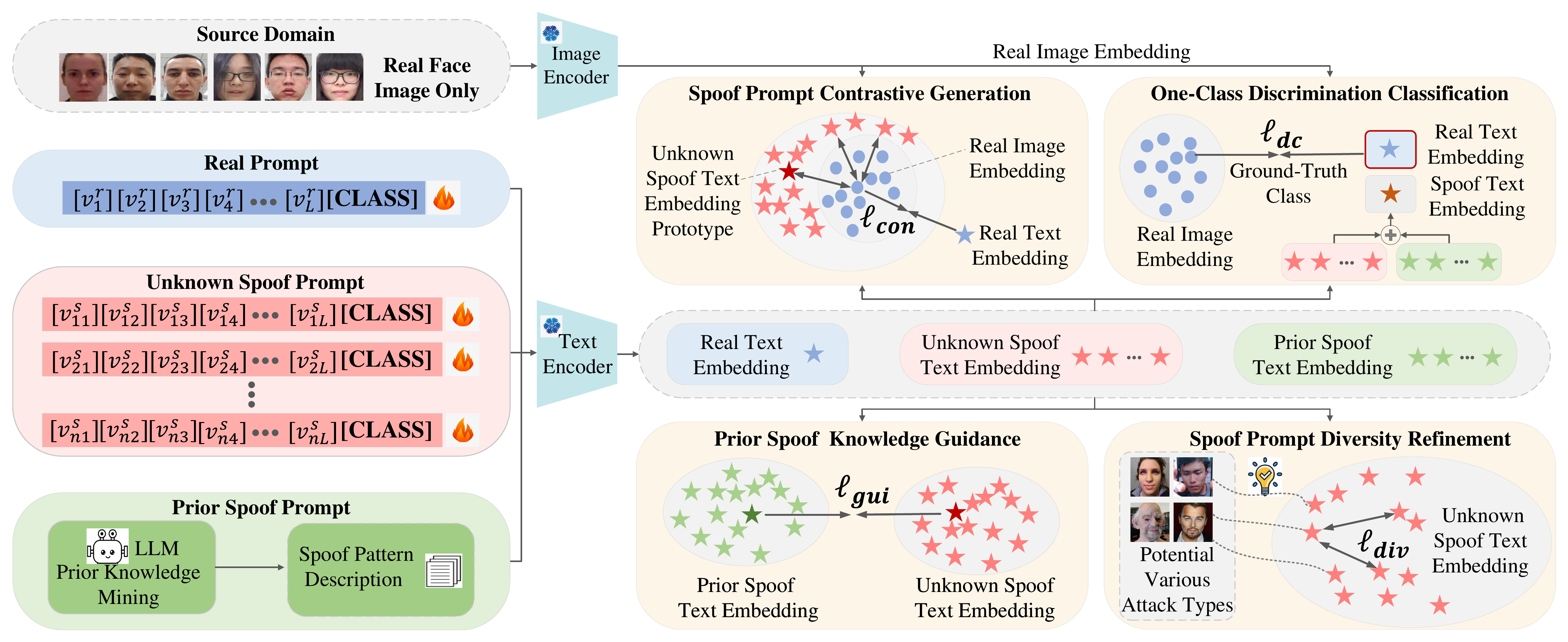}
	\caption{Overview of the unknown spoof prompt learning framework for generalized face anti-spoofing. 
This framework is centered on real face training images, iteratively extrapolating within the embedding space to generate unknown spoof prompts for potential attack types.  
To ensure the effectiveness of the learned prompts, four modules are proposed:  
a spoof prompt contrastive generation module, which ensures that the distance between unknown spoof prompts and real face images exceeds that between real prompts and real face images;  a prior spoof knowledge guidance module, which constrains unknown spoof prompts within the sparse range of prior spoof knowledge; a spoof prompt diversity refinement module,  which enforces semantic independence among different unknown spoof prompts;
a one-class discrimination module, which minimizes the discrepancy between the predicted probability distribution of real images and the ground-truth label distribution.
	}
	\label{fig2}
\end{figure*}

\section{Proposed Method}
\label{sec3}
\subsection{Problem Definition}
Let the single source domain, which contains only real face images, be denoted as $\mathcal{D}^s=\left\{x_{i}\right\}_{i=1}^{N}$, where $x_{i}$ represents the $i$-th training sample. Our task is to train a face anti-spoofing model using $\mathcal{D}^s$ such that it generalizes effectively to an unseen target domain  $\mathcal{D}^{m}=\left\{x_{i}^{m}\right\}_{i=1}^{N^{m}}$,  which exhibits both covariate shift and semantic shift relative to the source domain.
Here, $x_{i}^m$ represents the $i$-th test sample, which corresponds to either a real face or an unknown attack image.
This paper aims to learn textual prompts for real faces and potential unknown attack types, effectively leveraging the pre-trained general knowledge of vision-language models. The key challenge is deriving effective prompts for unknown attack types using only real faces.
To address this, we propose an unknown spoof prompt learning framework, as illustrated in \figurename~\ref{fig2}. The framework consists of a spoof prompt contrastive generation module, a spoof prompt diversity refinement module, a prior spoof knowledge guidance module, and a one-class discriminative classification module.

\subsection{Spoof Prompt Contrastive Generation}

A vision-language model usually consists of an image encoder $E_v$ and a text encoder $E_t$. Given an image $x_{i}$, its category is typically determined by identifying the textual prompt embedding that exhibits the highest similarity to its image embedding:
\begin{equation}\label{equ1}
	\begin{aligned}
{ \underset{k \in \left\{ {1,\ldots,K} \right\}}{{\arg\min} }  \frac{\exp\left(\rm{sim}\left(f_{v},f_t^k\right)/\tau\right)}{\sum_{k^{\prime}=1}^K\exp\left(\rm{sim}\left(f_{v},f_t^{k^{\prime}}\right)/\tau\right)}},
	\end{aligned}
\end{equation}
where $f_{v}=E_v(x_i)$ denotes the image embedding of  $x_i$, $f_t^k=E_t(t_k)$ represents
the embedding of the textual prompt $t_k$ for the $k$-th class, $K$ is the number of classes, $\rm{sim}$ is the similarity evaluation function, and $\tau$ is the temperature coefficient.
One basic assumption of this paper is that high-fidelity spoof faces tend to exhibit distributions in the embedding space that closely resemble those of real faces. 
Thus, expanding outward from real faces in the embedding space also facilitates the inference of potential textual prompts for spoof faces.  

Following the paradigm of typical prompt learning methods, such as CoOp~\citep{Zhou_2022}, we represent real and spoof prompts as learnable prompt vectors. Given the diversity of unknown attack types, which differ in material properties, three-dimensional structures, and color-texture characteristics, a single spoof prompt is insufficient to comprehensively capture the contextual information of all possible attacks. To address this, we assign individual spoof prompts to attack types exhibiting similar spoof patterns and aggregate these prompts into a set to better represent diverse unknown attacks.  
Based on this design, the real prompt $t^r$ and the set of unknown spoof prompts $t^u$ are defined as:
\begin{equation}\label{equ2}
	\begin{aligned}
		&t^r=[v_1^r][v_2^r]\ldots[v_L^r][CLASS],\\
		&t^u=\left\{   t^u_i|t^u_i=[v_{i1}^u][v_{i2}^u]\ldots[v_{iL}^u][CLASS],i \in \left\{ {1,\ldots,N^u} \right\}   \right\},
	\end{aligned}
\end{equation} 
where $[v_j^r]$ and $[v_{ij}^u]$ ($j \in \{1,2,\ldots,L\}$) denote the $j$-th context vectors, $L$ is the number of context tokens, and $N^u$ is the number of individual spoof prompts for unknown attacks.
Since the names of unknown attack types are unpredictable, we depart from previous approaches that assign distinct class names to different prompt types and instead unify the class names of $t^r$ and $t^u$ as ``human face". During prompt learning, we freeze the image encoder $E_v$ and text encoder $E_t$, optimizing only the prompt vectors. This facilitates the effective transfer of pre-trained vision-language model knowledge to the generalized face anti-spoofing task.

To ensure the effectiveness of both real and spoof face prompts, we propose a spoof prompt contrastive generation regularization term, $\ell_{con}$, to constrain the optimization process. $\ell_{con}$ is centered on real face image embeddings, maximizing the distance between spoof prompt embeddings and real face image embeddings while minimizing the distance between real prompt embeddings and real face image embeddings. For a given image $x_i \sim \mathcal{D}^s$, $\ell_{con}$ is formulated as:  
\begin{equation}\label{equ3}
	\begin{aligned}
		\ell_{con}=max\left(\lVert E_v(x_i)- E_t(t^r) \rVert_2  -  \lVert E_v(x_i)- E_t(t^{u\prime}) \rVert_2+\eta,0\right),
	\end{aligned}
\end{equation} 
where $t^{u\prime}$ represents the prototype of unknown spoof prompts, computed as the mean of all unknown spoof prompts.
The margin coefficient $\eta$ is set to $2$. By enforcing this contrastive regularization, the generated prompts for real and spoof faces become highly discriminative, while the extrapolated spoof prompts are expected to capture potential unknown attacks.

\subsection{Spoof Prompt Diversity Refinement}

The spoof prompt contrastive generation module incorporates contextual information about potential unknown attacks into spoof prompts from a global perspective. However, due to the diversity of unknown attack types, a key challenge remains: global optimization alone struggles to ensure that each individual spoof prompt captures fine-grained spoof patterns and that different spoof prompts correspond to distinct attack types.
To address this, we propose a diverse spoof prompt regularization term, $\ell_{div}$, to further refine the learnable spoof prompts and enhance their diversity. $\ell_{div}$ is defined as:
\begin{equation}\label{equ4}
	\begin{aligned}
		\ell_{div}=\sum_{i=1}^{N^u}\sum_{j\neq n}^{N^u}\rm{sim}(E_t(t_i^u),E_t(t_j^u)),
	\end{aligned}
\end{equation}
where $\rm{sim}$ denotes the similarity evaluation function, implemented as cosine similarity in this paper.

The regularization term $\ell_{div}$ encourages low similarity among spoof prompt vectors in the embedding space, thereby maximizing their semantic separation and ensuring their independence.  This promotes diversity among generated spoof prompts, increasing the likelihood that they correspond to distinct types of unknown attacks and encapsulate a broader range of spoof patterns. 

\subsection{Prior Spoof Knowledge Guidance}

During the outward expansion of spoof prompts, there is a risk of significant deviation from the embeddings of unknown attack images. To maintain the alignment of spoof prompts with realistic face anti-spoofing scenarios, we introduce a prior spoof-guided regularization term, denoted as $\ell_{gui}$, designed to constrain the unknown spoof prompts within a reasonable range.

Considering that large language models, trained on extensive datasets, inherently possess substantial knowledge regarding face anti-spoofing, we leverage this prior knowledge by formulating targeted queries to such models. Specifically, we construct precise prompt questions, such as, ``What are the typical anomaly patterns of spoof face images compared to real ones in face anti-spoofing?" and ``What are the key indicators for identifying spoof face images in face anti-spoofing?" to query ChatGPT~\citep{achiam2023gpt}. The obtained responses are manually curated and integrated to form descriptive texts called prior spoof prompts: $t^{p}=\{d_i| i \in \{1,2,\ldots,N^p\}\}$, where $d_i$ denotes an individual descriptive text, and $N^p$ represents the total number of texts.
The regularization term $\ell_{gui}$ is formally defined as:  
\begin{equation}\label{equ5}
	\begin{aligned}
		\ell_{gui}=\lVert E_t(t^{p\prime})- E_t(t^{u\prime}) \rVert_2 ,
	\end{aligned}
\end{equation} 
where $t^{p\prime}$ is the prototype of prior spoof prompts.
$\ell_{gui}$ minimizes the embedding distance between unknown spoof prompts and prior spoof prompts. This facilitates the acquisition of meaningful contextual information related to plausible unknown attack types.
Finally, we concatenate the unknown spoof prompts $t^{u}$ with the prior spoof prompts $t^{p}$, and use the prototype of the combined prompts as the overall spoof prompt $t^{s}$.

\subsection{One-Class Discriminative Classification}
Given that the source domain comprises exclusively real face images, we propose a one-class discriminative classification regularization term, denoted as $\ell_{dc}$, to minimize the discrepancy between the predicted probability distribution of real images and the ground-truth label distribution. The ground-truth labels are defined across two classes: real and spoof faces.  The predicted probability distribution is computed based on the similarity between image embeddings and embeddings of real or spoof text prompts.

Formally, for a given image $x_i \sim \mathcal{D}^s$,  $\ell_{dc}$ is implemented as a cross-entropy loss:
\begin{equation}\label{equ6}
	\begin{aligned}
		\ell_{dc}=-\log \frac{\exp\left(\rm{sim}\left(E_{v}(x_i),E_t(t^r)\right)/\tau\right)}{\sum\limits_{t^{k} \in \left\{ {t^r,t^s} \right\}}\exp\left(\rm{sim}\left(E_{v}(x_i),E_t(t^{k}) \right)/\tau\right)}.
	\end{aligned}
\end{equation} 
The regularization term $\ell_{dc}$ explicitly enhances the discriminative ability of real prompts towards real face images while implicitly reducing the similarity between real images and spoof prompts.

\subsection{Overall Training Object and Inference}
Overall, the optimization objective for learning real and spoof prompts is expressed as:
\begin{equation}\label{equ7}
	\begin{aligned}
		\ell_{all}=\lambda_1\ell_{dc}+\lambda_2\ell_{con}+\lambda_3\ell_{div}+\lambda_4\ell_{gui},
	\end{aligned}
\end{equation}
where $\lambda_1$, $\lambda_2$, $\lambda_3$, $\lambda_4$ are hyper-parameters used to balance the contributions of each loss component.
The synergistic constraints introduced by these four regularization terms facilitate the development of highly representative and diverse fine-grained textual prompts for both real faces and unknown attack types. Consequently, this enhances the transferability of visual-language model knowledge, enabling more effective handling of covariate and semantic shifts in unknown target domains.

During inference, we predict the probability distribution of a given test sample using the learned real prompts $t^r$ and spoof prompts $t^s$ following the methodology described in Equation~\ref{equ1}. A conventional threshold-based decision approach is then applied to classify the test sample as either a real or spoof face.

\section{Experiments}
\label{sec4}
\subsection{Experimental Setups}
\label{sec41}
\begin{figure}[!t]
	\centering
	\includegraphics[width=0.5\textwidth]{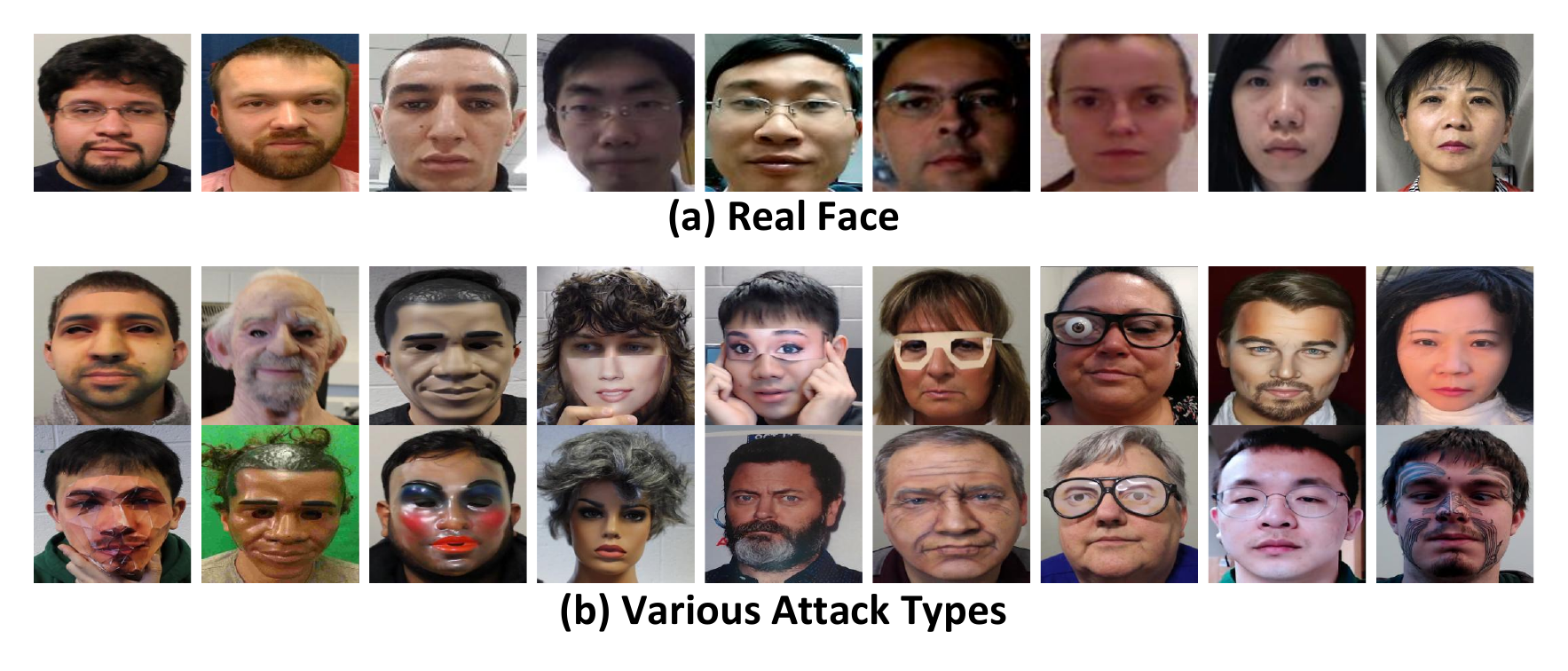}
	\caption{Sample images from the nine datasets. (a) Due to variations in recording equipment and environments, the real face images across the nine datasets exhibit significant covariate shift in their visual characteristics.  
		(b) In addition to covariate shift, the diversity of attack types results in significant semantic shift due to variations in spoof patterns across different attack types.
	}
	\label{fig7}
\end{figure}

\noindent \textbf{Datasets.}
We evaluate the performance of the proposed method on nine commonly used datasets: SiW-Mv2~\citep{guo2022multi}, WMCA~\citep{george2019biometric}, CASIA-MFSD~\citep{zhang2012face} (C for short), Replay-Attack~\citep{chingovska2012effectiveness} (I), MSU-MFSD~\citep{wen2015face} (M), OULU-NPU~\citep{boulkenafet2017oulu} (O), 3DMAD~\citep{erdogmus2014spoofing2} (D), HKBU-MARs~\citep{liu20163dr} (H),
CASIA-SURF 3DMask~\citep{yu2020fas} (U). Similar to previous approaches, we treat each dataset as a distinct domain. 
Sample images from these datasets are shown in \figurename~\ref{fig7}.

These datasets exhibit significant variations in recording environments, capture devices, lighting conditions, manufacturing techniques, and attack types.
The O, C, M, I four datasets include only two types of attacks: print and replay. The D, H, U three datasets contain 3D masks made from different materials, such as resin and silicone.
In contrast, the SiW-Mv2 and WMCA datasets feature a broad range of attack types. Specifically, the SiW-Mv2 dataset includes 14 different attack types: half mask, paper mask, transparent mask, silicone mask, mannequin, partial eye, funny eye glasses, partial mouth, paper glasses, replay, print, cosmetic makeup, impersonation makeup, obfuscation makeup, while the WMCA dataset contains 7 different attack types: fake head, paper mask, rigid mask, flexible mask, glasses, replay, print. 
For clarity and conciseness in tables, the attack types of the SiW-Mv2 dataset are abbreviated as 
Hal., Pap., Tra., Sil., Man., Eye, Fun., Mou., Gla., Rep., Pri., Cos., Imp., Obf., respectively.

\noindent \textbf{Evaluation protocols and metrics.}
Due to the substantial covariate and semantic disparities between the source and unseen target domains, following OC-SCMNet~\citep{huang2024one}, we utilize the O, M, C, I, D, H, U datasets to construct six cross-dataset, cross-attack-type evaluation protocols, as shown in Table~\ref{tab1}.
These protocols are specifically designed to assess cross-domain performance across print, replay, and 3D mask attack types.

Considering the existence of numerous additional attack types, we utilize the publicly available SiW-Mv2 and WMCA datasets, which cover a wider range of attack types, to design more comprehensive cross-dataset and cross-attack-type evaluation protocols. Due to the extensive variety of attack types, we adopt a leave-one-attack-out strategy to evaluate the model’s generalization ability to specific unknown attack types. As presented in Table~\ref{tab1}, in each protocol, the images of the withheld type of attack, together with the images of the real faces, constitute the unseen target domain. 
Unless otherwise specified, only real face images from the source domain are used during training, while both real and spoof face images from the target domain are used for testing.

\begin{table}[!tbp]
	\footnotesize
	\caption{All the 27 evaluation protocols. In performance evaluation experiments of our method, when multiple datasets are available, they are combined to form a single domain.}
	\label{tab1}
	\centering 
	\begin{tabular}{|p{1.3cm}<{\centering}|p{2.2cm}|p{3.6cm}|
		}
		\hline
		Protocol& Source Domain&Unseen Target Domain  \\
        \hline
		P1 &OM (real)&DHU (real and 3D mask)\\
        \hline
		P2 &OMCI (real)&DHU (real and 3D mask)\\
        \hline
		P3 &OMD (real)&OMCI (real and print)\\
        \hline
		P4 &OMCIDHU (real)&OMCI (real and print)\\
        \hline
		P5 &OMD (real)&OMCI (real and replay)\\
        \hline
		P6 &OMCIDHU (real)&OMCI (real and replay)\\
		\hline
        A1-A14 & WMCA (real)&SiW-Mv2 (real and a single withheld attack type from the 14 attack types)\\
		\hline
        B1-B7 & SiW-Mv2 (real)&WMCA (real and  a single withheld attack type from the 7 attack types)\\
		\hline
    \end{tabular}
\end{table}

We utilize several widely adopted evaluation metrics to assess the performance of the proposed method in comparison to existing face anti-spoofing approaches. These metrics include the Attack Presentation Classification Error Rate (APCER), Bonafide Presentation Classification Error Rate (BPCER), Average Classification Error Rate (ACER), Area Under Curve (AUC), and Half Total Error Rate (HTER).

\begin{table*}[!t]
	\footnotesize
	\caption{Cross-domain leave-one-attack-out evaluation results using the WMCA dataset as the source domain and the SiW-Mv2 dataset as the unseen target domain. Red and blue labels represent the best-performing and second-best-performing in terms of ACER.} 
	\label{tabcross1}
	\centering 
	\begin{tabular}{|p{1.5cm}<{\centering}|p{1.1cm}<{\centering}|p{0.6cm}<{\centering}|p{0.6cm}<{\centering}|p{0.6cm}<{\centering}|p{0.6cm}<{\centering}|p{0.6cm}<{\centering}|p{0.6cm}<{\centering}|p{0.6cm}<{\centering}|p{0.6cm}<{\centering}|p{0.6cm}<{\centering}|p{0.6cm}<{\centering}|p{0.6cm}<{\centering}|p{0.6cm}<{\centering}|p{0.6cm}<{\centering}|p{0.6cm}<{\centering}|p{0.6cm}<{\centering}|
		}
		\hline	
		\multirow{2}{*}{Method}&\multirow{2}{*}{Metric(\%)}&\multicolumn{5}{c|}{ Mask Attacks}& \multicolumn{4}{c|}{Partial Attacks}&\multicolumn{2}{c|}{2D Attacks}&\multicolumn{3}{c|}{Makeup Attacks}& \multirow{2}{*}{Avg.} \\ \cline{3-16}
		            &&Hal. & Pap.&Tra. &Sil. & Man.&Eye&Fun.&  Mou.&Gla.&Rep.&Print&Cos.&Imp.& Obf.&     \\ \hline

\multirow{4}{*}{\parbox{1.5cm}{\centering OC-CNN \\~\citep{oza2018one}}}&APCER$\downarrow$&0.00&99.61&82.38&0.00&87.74&99.62&0.77&0.00&2.68&99.23&5.36&0.38&10.73&0.00& 34.89        \\ 
&BPCER$\downarrow$&100.00&0.00&8.33&100.00&5.00&0.00&100.00&100.00&100.00&2.04&63.70&100.00&86.89&100.00&61.85         \\ 
&ACER$\downarrow$&50.00&49.81&45.35&50.00&46.37&49.81&50.38&50.00&51.34&50.64&34.53&50.19&48.80&50.00&48.37         \\ 
&AUC$\uparrow$&51.28&66.54&50.79&66.24&74.66&59.84&30.14&76.44&23.02&50.46&71.74&34.94&48.38&57.04& 54.39        \\ \hline
		
\multirow{4}{*}{\parbox{1.5cm}{\centering AD \\~\citep{baweja2020anomaly}}}&APCER$\downarrow$&54.79&43.30&100.00&100.00&39.08&68.97&100.00&100.00&55.55&54.41&32.57&59.00&54.41&61.30& 65.96        \\ 
&BPCER$\downarrow$&52.78&41.18&0.00&0.00&40.00&70.18&0.00&0.00&48.68&47.96&34.81&55.77&59.02&59.09& 36.39        \\ 
&ACER$\downarrow$&53.78&42.24&50.00&50.00&39.54&69.57&50.00&50.00&52.12&51.18&33.69&57.39&56.71&60.20& 51.17        \\ 
&AUC$\uparrow$&45.96&58.03&46.11&59.22&63.80&25.66&39.86&59.37&44.49&48.26&69.19&37.94&45.07&32.10& 48.21        \\ \hline

\multirow{4}{*}{\parbox{1.5cm}{\centering Hyp-OC\\ ~\citep{narayan2024hyp}}}&APCER$\downarrow$&37.93&13.41&49.81&19.15&30.65&27.20&50.57&18.39&45.59&21.07&33.72&24.52&38.31&21.84&31.00         \\ 
&BPCER$\downarrow$&33.94&11.76&43.33&17.64&10.00&5.26&33.52&17.24&25.00&22.45&34.07&32.69&24.59&18.18& 23.13        \\ 
&ACER$\downarrow$&\textcolor{blue}{35.94}&\textcolor{blue}{12.59}&46.57&18.40&20.33&\textcolor{blue}{16.23}&\textcolor{blue}{42.05}&\textcolor{blue}{17.82}&35.30&\textcolor{red}{21.76}&33.90&28.61&31.45&\textcolor{red}{20.01}&27.07         \\ 
&AUC$\uparrow$&63.20&92.99&52.95&82.71&82.07&86.51&57.75&87.55&67.00&85.35&67.25&70.71&75.81&79.47& 76.05        \\ \hline
		
\multirow{4}{*}{\parbox{1.5cm}{\centering OC-SCMNet\\~\citep{huang2024one}}}&APCER$\downarrow$&40.28&0.01&10.00&0.01&27.50&1.75&16.76&3.45&38.16&50.00&9.63&19.23&6.56&13.64& 16.92        \\ 
&BPCER$\downarrow$&35.63&40.99&24.90&35.63&11.49&48.28&70.88&45.97&25.29&11.88&42.53&37.93&13.79&52.49&35.55         \\	
&ACER$\downarrow$&37.95&20.50&\textcolor{blue}{17.45}&\textcolor{blue}{17.82}&\textcolor{blue}{19.50}&25.02&43.82&24.71&\textcolor{blue}{31.72}&30.94&\textcolor{blue}{26.08}&\textcolor{blue}{28.58}&\textcolor{blue}{10.18}&\textcolor{blue}{33.06}& \textcolor{blue}{26.23}        \\	
&AUC$\uparrow$&61.79&68.92&87.59&85.67&85.07&71.97&54.09&74.94&70.14&68.75&75.02&71.04&94.49&58.10&   73.40      \\	\hline

\multirow{4}{*}{\parbox{1.5cm}{\centering Ours}}&APCER$\downarrow$&47.89&8.81&16.86&9.58&3.07&17.24&40.23&13.03&36.78&21.46&9.20&35.10&3.06&12.64&19.64\\ 
&BPCER$\downarrow$&22.22&0.00&8.33&0.00&0.00&1.75&40.78&6.90&25.00&33.67&28.89&21.15&1.64&63.64&18.14\\	
&ACER$\downarrow$&\textcolor{red}{35.06}&\textcolor{red}{4.40}&\textcolor{red}{12.60}&\textcolor{red}{4.79}&\textcolor{red}{1.53}&\textcolor{red}{9.50}&\textcolor{red}{40.51}&\textcolor{red}{9.96}&\textcolor{red}{30.89}&\textcolor{blue}{27.56}&\textcolor{red}{19.04}&\textcolor{red}{28.13}&\textcolor{red}{2.35}&38.14&\textcolor{red}{18.89}\\ 
&AUC$\uparrow$&65.68&95.33&92.71&98.47&99.42&95.17&58.88&94.81&70.67&76.81&84.49&71.45&99.19&54.52&82.69\\				\hline
	\end{tabular}
\end{table*}

\noindent \textbf{Competitors.}
We compare the proposed method with previous state-of-the-art one-class face anti-spoofing methods: OC-CNN~\citep{oza2018one}, IQM-GMM~\citep{nikisins2018effectiveness}, AD~\citep{baweja2020anomaly},  AAE~\citep{huang2021one}, Hyp-OC~\citep{narayan2024hyp}, OC-SCMNet~\citep{huang2024one}.
OC-CNN and IQM-GMM are classic approaches that train one-class classifiers for face anti-spoofing. In contrast, AD, AAE, Hyp-OC, and OC-SCMNet take a more proactive approach by generating pseudo-spoof patterns during training to enhance the discriminability between real and spoof faces. Notably, recent methods such as Hyp-OC and OC-SCMNet have achieved significant improvements in cross-scenario generalization performance.

\noindent \textbf{Implementation details.}
The proposed method is implemented using PyTorch. A face detector~\cite{Zhang2016Joint} is employed to detect and align face images from raw video frames. The pre-trained vision-language model is based on the CLIP model with a ViT-B/16 backbone. For prompt learning, stochastic gradient descent (SGD) with a momentum of 0.9 and a weight decay of 0.0005 is used as the optimizer. The learning rate is initially set to 0.02 and updated according to a cosine annealing schedule. The batch size is set to 64. The hyper-parameters $\lambda_1$, $\lambda_2$, $\lambda_3$, $\lambda_4$ are set to 0.5, 1, 1, and 1, respectively. The number of individual spoof prompts $N^u$ is set to 12.

\begin{table*}[!tbp]
	\footnotesize
	\caption{Cross-domain leave-one-attack-out evaluation results using the SiW-Mv2 dataset as the source domain and the WMCA dataset as the unseen target domain. Red and blue labels represent the best-performing and second-best-performing in terms of ACER.}
	\label{tabcross2}
	\centering 
	\begin{tabular}{|p{4.25cm}|p{1.15cm}<{\centering}|p{1.15cm}<{\centering}|p{1.15cm}<{\centering}|p{1.15cm}<{\centering}|p{1.15cm}<{\centering}|p{1.15cm}<{\centering}|p{1.15cm}<{\centering}|p{1.15cm}<{\centering}|p{1.2cm}<{\centering}|
		}
		\hline
		Method& Metric (\%)  & Fake Head  & Paper Mask  & Rigid Mask & Flexible Mask  & Glasses & Replay& Print&Average\\ \hline
		
		\multirow{4}{*}{OC-CNN~\citep{oza2018one}}
		&APCER$\downarrow$&93.91&0.00&0.87&0.00&2.61&57.39&0.00&22.11         \\
		&BPCER$\downarrow$&16.39&100.00&100.00&100.00&100.00&34.60&100.00&78.71         \\
		&ACER$\downarrow$&55.15&50.00&50.43&50.00&51.30&45.99&50.00&50.41         \\
		&AUC$\uparrow$&33.33&48.39&50.72&45.41&38.63&51.40&53.42& 45.90        \\
		
		\hline 
		\multirow{4}{*}{AD~\citep{baweja2020anomaly}}&
		APCER$\downarrow$&100.00&100.00&42.61&45.22&100.00&100.00&100.00&83.98         \\
		&BPCER$\downarrow$&0.00&0.00&42.86&41.42&0.00&0.00&0.00&12.04         \\
		&ACER$\downarrow$&50.00&50.00&42.73&43.32&50.00&50.00&50.00&48.00         \\
		&AUC$\uparrow$&43.17&45.36&53.28&60.57&56.15&38.66&31.46&46.95         \\
		
		\hline
		\multirow{4}{*}{Hyp-OC~\citep{narayan2024hyp}}
		&APCER$\downarrow$&19.13&29.57&40.00&37.39&33.04&13.91&23.48&28.07         \\
		&BPCER$\downarrow$&17.21&40.85&36.19&34.56&34.58&14.53&22.78& 28.67        \\
		&ACER$\downarrow$&18.17&35.21&38.10&35.98&33.81&\textcolor{red}{14.22}&23.13& 28.37        \\
		&AUC$\uparrow$&85.36&68.53&65.56&65.60&72.36&91.43&83.88& 76.10        \\
		
		\hline
		\multirow{4}{*}{OC-SCMNet~\citep{huang2024one}}
		&APCER$\downarrow$&29.51&14.08&55.24&40.11&43.93&25.95&19.31&32.59         \\				
		&BPCER$\downarrow$&5.22&24.34&4.35&12.17&2.61&3.48&0.87&  7.58       \\				
		&ACER$\downarrow$&\textcolor{blue}{17.36}&\textcolor{blue}{19.22}&\textcolor{blue}{29.79}&\textcolor{blue}{26.14}&\textcolor{blue}{23.27}&\textcolor{blue}{14.71}&\textcolor{blue}{10.09}& \textcolor{blue}{20.08}        \\				
		&AUC$\uparrow$&82.37&86.18&71.26&79.22&77.07&93.50&92.31&  83.13       \\	
		
		\hline
		\multirow{4}{*}{Ours}
		&APCER$\downarrow$&0.00&0.87&7.83&2.61&17.39&9.57&8.70&6.71\\
		&BPCER$\downarrow$&9.01&1.41&40.00&29.02&26.17&49.48&10.42&23.64\\	
		&ACER$\downarrow$&\textcolor{red}{4.51}&\textcolor{red}{1.14}&\textcolor{red}{23.91}&\textcolor{red}{15.82}&\textcolor{red}{21.78}&29.52&\textcolor{red}{9.56}&\textcolor{red}{15.18}\\		
		&AUC$\uparrow$&96.32&99.85&80.95&90.69&83.12&68.08&94.12&87.59\\			
		\hline
	\end{tabular}
\end{table*}

\begin{table*}[!tbp]
	\footnotesize
	\caption{Cross-domain leave-one-attack-out evaluation results on protocols P1-P6. Red and blue labels represent the best-performing and second-best-performing.}
	\label{tabcross3}
	\centering 
	\begin{tabular}{|p{4.25cm}|p{0.8cm}<{\centering}|p{0.71cm}<{\centering}|p{0.8cm}<{\centering}|p{0.71cm}<{\centering}|p{0.8cm}<{\centering}|p{0.71cm}<{\centering}|p{0.8cm}<{\centering}|p{0.71cm}<{\centering}|p{0.8cm}<{\centering}|p{0.71cm}<{\centering}|p{0.8cm}<{\centering}|p{0.71cm}<{\centering}|p{0.8cm}<{\centering}|p{0.8cm}<{\centering}|p{0.8cm}<{\centering}|
		}
		\hline
		\multirow{2}{*}{Method}&  \multicolumn{2}{c|}{P1} & \multicolumn{2}{c|}{P2} & \multicolumn{2}{c|}{P3} & \multicolumn{2}{c|}{P4} & \multicolumn{2}{c|}{P5}& \multicolumn{2}{c|}{P6} \\ \cline{2-13}
		&HTER$\downarrow$&AUC$\uparrow$&HTER$\downarrow$&AUC$\uparrow$&HTER$\downarrow$&AUC$\uparrow$&HTER$\downarrow$&AUC$\uparrow$&HTER$\downarrow$&AUC$\uparrow$&HTER$\downarrow$&AUC$\uparrow$\\ \hline
		IQM-GMM~\citep{nikisins2018effectiveness}&43.58& 46.99& 43.82& 47.18& 40.25& 62.02& 47.56& 41.68& 37.61 &\textcolor{blue}{64.66} &48.78& 41.85\\ \hline
		AAE~\citep{huang2021one}&42.85& 55.97& 41.07& 55.35& 48.50& 40.94& 42.69& 57.21 &46.70& 53.94& 37.60& 64.68\\ \hline
		AD~\citep{baweja2020anomaly}&39.35& 61.86& 42.19& 57.47& 41.59& 61.56& 40.41& 63.83& 48.06& 42.45& 46.87& 41.26\\ \hline
		Hyp-OC~\citep{narayan2024hyp}&47.47&53.64&47.93&57.85&41.05&61.13&43.80&57.17&40.35&64.22&36.79&68.34\\ \hline
		OC-SCMNet~\citep{huang2024one}&\textcolor{blue}{24.14}& \textcolor{blue}{74.81}& \textcolor{red}{20.85}& \textcolor{red}{85.40}& \textcolor{blue}{37.44}& \textcolor{blue}{63.23}& \textcolor{blue}{28.99}& \textcolor{blue}{72.21} &\textcolor{blue}{36.41}& 63.56& \textcolor{blue}{29.61}& \textcolor{blue}{74.99}  \\ \hline	
		Ours&\textcolor{red}{22.04}&\textcolor{red}{82.02}&\textcolor{blue}{20.88}&\textcolor{blue}{80.97}&\textcolor{red}{22.96}&\textcolor{red}{84.16}&\textcolor{red}{15.44}&\textcolor{red}{91.64}&\textcolor{red}{22.60}&\textcolor{red}{85.35}&\textcolor{red}{3.76}&\textcolor{red}{99.36}\\ 
		\hline
	\end{tabular}
\end{table*}

\subsection{Comparison with State-of-the-Art Face Anti-Spoofing Methods}
\textbf{Cross-Domain Leave-One-Attack-Out Evaluation.}
We compare our method with previous state-of-the-art face anti-spoofing methods using cross-domain leave-one-attack-out protocols in unseen target domains. The experimental results are presented in Tables~\ref{tabcross1}, Tables~\ref{tabcross2}, and Tables~\ref{tabcross3}.  

Early methods~\citep{oza2018one,baweja2020anomaly} that rely on the distribution of real face images are sensitive to covariate and semantic shifts, leading to poor generalization across scenarios. In contrast, recent approaches~\citep{narayan2024hyp,huang2024one} have significantly improved the generalization capability of face anti-spoofing models for unseen target domains by leveraging hyperbolic space mapping and generating spoof cues independent of the real face data distribution.  

\begin{figure}[!tbp]
	\centering
	\includegraphics[width=0.45\textwidth]{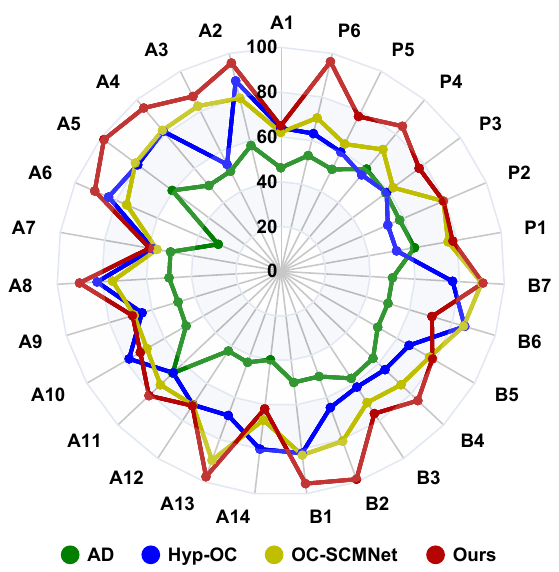}
	\caption{Overall performance comparison across all the 27 protocols. (100-ACER)$\uparrow$(\%) on protocols A1-A14 and B1-B7. (100-HTER)$\uparrow$(\%) on on protocols P1-P6.
	}
	\label{fig8}
\end{figure}

Compared to recent state-of-the-art methods~\citep{narayan2024hyp,huang2024one}, our approach achieves substantial performance improvements in 23 out of the 27 evaluation protocols. For instance, in the fourteen protocols where the SiW-Mv2 dataset serves as the target domain, our method reduces the average ACER by 27.98\% compared to OC-SCMNet~\citep{huang2024one}. In the seven protocols where the WMCA dataset serves as the target domain, the average ACER is reduced by 24.40\%. While in the P1–P6 protocols, our approach lowers the average HTER by 39.31\%. Similarly, the AUC metric also demonstrates the significant performance improvement achieved by our method.  These substantial improvements indicate that our approach, which relies solely on real face images, effectively learns representative real and spoof prompts for real faces and potential unknown attack types.

From the perspective of specific attack types, our method demonstrates particularly notable performance improvements for print, impersonation makeup, paper mask, transparent mask, silicone mask, mannequin, partial eye, funny eye glasses, and partial mouth attacks. For instance, in terms of the ACER values in Tables~\ref{tabcross1}, impersonation makeup is improved from 10.18\% to 2.35\%, while silicone mask is enhanced from 17.82\% to 4.79\%. 
The performance improvements across various attack types not only demonstrate the effectiveness of our method in detecting high-quality unknown attacks but also underscore its capability to generalize across a diverse range of potential attack types.
This indicates that the learned spoof prompts exhibit diverse spoof patterns, providing the vision-language model with effective contextual information for improved detection.

\begin{table*}[!tbp]
	\footnotesize
	\caption{Comparison with the classic prompt learning method CoOp under cross-domain leave-one-attack-out protocols using the WMCA dataset as the source domain and the SiW-Mv2 dataset as the unseen target domain. Red and blue labels represent the best-performing and second-best-performing in terms of ACER.}
	\label{tabcoop}
	\centering 
	\begin{tabular}{|p{1.4cm}<{\centering}|p{1.1cm}<{\centering}|p{0.6cm}<{\centering}|p{0.6cm}<{\centering}|p{0.6cm}<{\centering}|p{0.6cm}<{\centering}|p{0.6cm}<{\centering}|p{0.6cm}<{\centering}|p{0.6cm}<{\centering}|p{0.6cm}<{\centering}|p{0.6cm}<{\centering}|p{0.6cm}<{\centering}|p{0.6cm}<{\centering}|p{0.6cm}<{\centering}|p{0.6cm}<{\centering}|p{0.6cm}<{\centering}|p{0.6cm}<{\centering}|
		}
		\hline	
		\multirow{2}{*}{Method}&\multirow{2}{*}{Metric(\%)}&\multicolumn{5}{c|}{ Mask Attacks}& \multicolumn{4}{c|}{Partial Attacks}&\multicolumn{2}{c|}{2D Attacks}&\multicolumn{3}{c|}{Makeup Attacks}& \multirow{2}{*}{Avg.} \\ \cline{3-16}
		            &&Hal. & Pap.&Tra. &Sil. & Man.&Eye&Fun.&  Mou.&Gla.&Rep.&Print&Cos.&Imp.& Obf.&     \\ \hline		

	\multirow{4}{*}{\parbox{1.5cm}{\centering CoOp\\~\citep{Zhou_2022}}}
        &APCER$\downarrow$&45.98&32.57&41.00&68.20&40.23&26.44&41.00&26.44&43.30&27.97&36.78&26.44&26.82&37.93&37.22\\ 
		&BPCER$\downarrow$&45.83&0.00&38.33&23.53&10.00&8.77&42.46&3.45&44.74&28.57&5.93&28.85&4.92&36.36&22.98\\	
		&ACER$\downarrow$&\textcolor{blue}{45.91}&\textcolor{blue}{16.28}&\textcolor{blue}{39.66}&\textcolor{blue}{45.86}&\textcolor{blue}{25.11}&\textcolor{blue}{17.60}&\textcolor{blue}{41.73}&\textcolor{blue}{14.94}&\textcolor{blue}{44.02}&\textcolor{blue}{28.27}&\textcolor{blue}{21.35}&\textcolor{red}{26.44}&\textcolor{blue}{15.87}&\textcolor{red}{37.15}&\textcolor{blue}{30.01}\\	
		&AUC$\uparrow$&55.03&83.72&62.68&52.13&76.59&84.21&61.04&86.13&58.82&77.19&79.02&77.97&85.46&66.93&71.92\\	
		\hline
		\multirow{4}{*}{Ours}&APCER$\downarrow$&47.89&8.81&16.86&9.58&3.07&17.24&40.23&13.03&36.78&21.46&9.20&35.10&3.06&12.64&19.64\\ 
		&BPCER$\downarrow$&22.22&0.00&8.33&0.00&0.00&1.75&40.78&6.90&25.00&33.67&28.89&21.15&1.64&63.64&18.14\\	
		&ACER$\downarrow$&\textcolor{red}{35.06}&\textcolor{red}{4.40}&\textcolor{red}{12.60}&\textcolor{red}{4.79}&\textcolor{red}{1.53}&\textcolor{red}{9.50}&\textcolor{red}{40.51}&\textcolor{red}{9.96}&\textcolor{red}{30.89}&\textcolor{red}{27.56}&\textcolor{red}{19.04}&\textcolor{blue}{28.13}&\textcolor{red}{2.35}&\textcolor{blue}{38.14}&\textcolor{red}{18.89}\\	
		&AUC$\uparrow$&65.68&95.33&92.71&98.47&99.42&95.17&58.88&94.81&70.67&76.81&84.49&71.45&99.19&54.52&82.69\\				\hline
		
	\end{tabular}
\end{table*}

\textbf{Overall Performance Visualization Analysis.}
We also use a radar chart to provide a comprehensive visualization of the performance of all approaches across all protocols. The results are presented in \figurename~\ref{fig8}.
The results indicate that our method exhibits superior generalization capabilities compared to existing approaches. This suggests that, despite not encountering spoof face images during training, our method effectively learns diverse and highly discriminative prompts for various potential attack types. The likely reason for this is that our method successfully adapts the pre-trained knowledge of the vision-language model to tackle the task of distinguishing between real and spoof faces.

\begin{figure}[!t]
	\centering
	\includegraphics[width=0.5\textwidth]{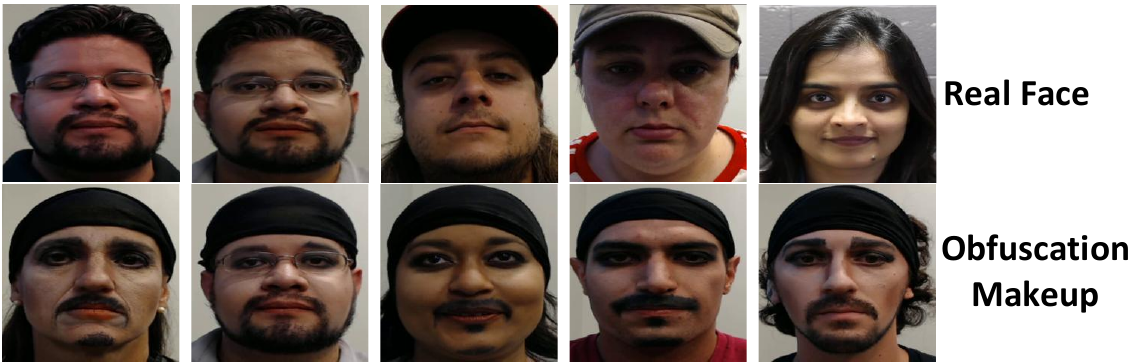}
	\caption{Challenging samples under the A14 obfuscation makeup protocol.
	}
	\label{fig9}
\end{figure}

As shown in \figurename~\ref{fig8}, our method performs less effectively than existing approaches under the obfuscation makeup protocol. This type of attack relies heavily on facial makeup to mimic the appearance of legitimate users. Some challenging samples from the A14 obfuscation makeup protocol are shown in \figurename~\ref{fig9}. Visually, real faces and faces with obfuscation makeup appear highly similar, with the primary differences stemming from changes in texture and surface reflectance due to the heavy cosmetic application. The suboptimal performance on the obfuscation makeup protocol suggests that our method still has room for improvement in capturing this specific type of spoof pattern.

\subsection{Comparison with Classic Prompt Learning Methods}
Our method builds upon the classic prompt learning approach, CoOp~\citep{Zhou_2022}, with optimized prompt vectors and learning processes specifically designed for the cross-scenario face anti-spoofing task. In this paper, we also compare our method with CoOp using cross-domain leave-one-attack-out protocols, where the WMCA dataset serves as the source domain and the SiW-Mv2 dataset as the unseen target domain. Notably, CoOp requires different classes of images to learn class-specific prompts. Thus, both real and spoof face images are used during CoOp training. Besides, some attack types in the SiW-Mv2 and WMCA datasets overlap, and during CoOp training, we exclude images of corresponding attack types from the source domain based on the attack type present in the target domain. CoOp initializes the category names as "real face" and "spoof face."

\textbf{Quantitative Analysis.} The results of the quantitative analysis are presented in Table~\ref{tabcoop}. As shown, despite CoOp utilizing additional spoof face images for training, our method still outperforms it in 12 out of the 14 protocols. A likely reason for CoOP's inferior generalization to diverse attack types is the inherent limitation of a single-category prompt vector, which provides constrained contextual information and is prone to overfitting the attack types seen during training. The elevated APCER further corroborates this observation. In contrast, our method optimizes the prompt vectors in the embedding space solely based on real face training images, effectively inferring potential attack-related prompts. 
This provides richer and more precise contextual information for adapting the general knowledge of the visual-language model to the face anti-spoofing task.
 
\begin{figure}[!t]
	\centering
	\includegraphics[width=0.5\textwidth]{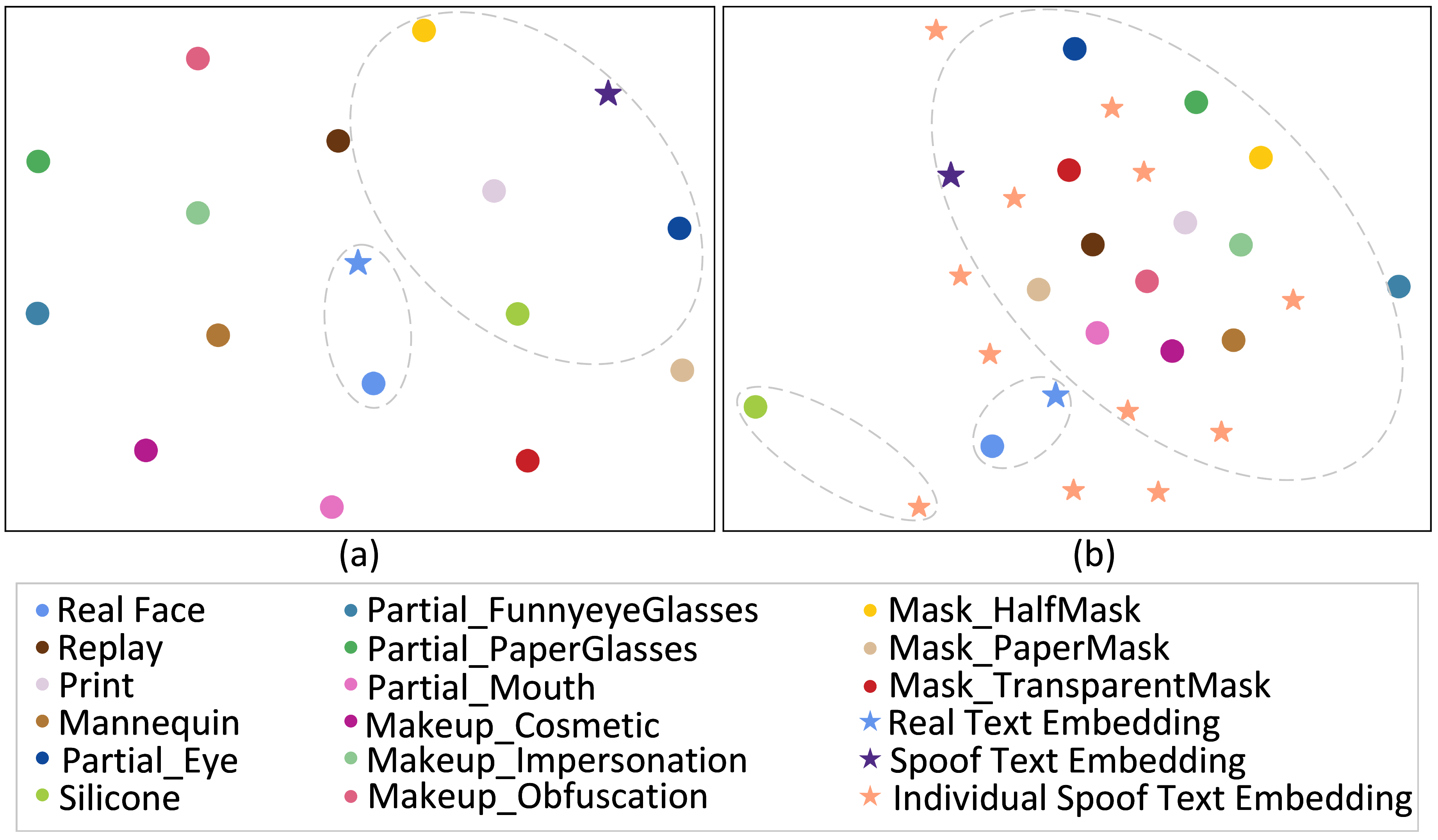}
	\caption{Image prototype and text prompts visualization for real face and various attack types on the protocol WMCA to SiW-Mv2 leaving the silicone mask out. (a) CoOp, (b) Ours.
	}
	\label{fig3}
\end{figure}

\textbf{Visualization Analysis.}
To further compare our method with CoOp, we visualize the prototypes of different images and their corresponding learned textual prompts in the embedding space. Since the silicone face mask is a critical attack type, we conduct a comparison on the protocol where WMCA serves as the source domain, and SiW-Mv2 as the target domain, where the silicone face mask is left as the attack type. The comparison results are shown in \figurename~\ref{fig3}.

In CoOp's embedding space, half of the attack image prototypes are closer to the real prompt embeddings than to the spoof prompt embeddings. This indicates that the spoof prompts learned by CoOp struggle to capture the characteristics of diverse attack types effectively, even though all attack types, except for the silicone mask, are included in the training data. As a result, the learned prompts exhibit poor discriminative ability between real and spoof faces.  
In contrast, our method, despite never encountering any attack images during training, learns spoof prompts that are well-distributed across the image embeddings of various unknown attack types. More attack image prototypes are closer to the spoof prompt embeddings than to the real prompt embeddings, indicating that the learned prompts are better at distinguishing real from spoof faces. This demonstrates that our method effectively generates meaningful textual prompts for both real faces and potential attack types. Furthermore, the dispersed distribution of spoof prompts underscores the effectiveness of the diverse spoof prompt regularization, ensuring the variability of the learned spoof prompts.

\begin{table}[!tbp]
	\footnotesize
	\caption{Component analysis results under the protocol WMCA to SiW-Mv2 leaving the silicone mask out. Unknown Spoof Prompt is short as USP and Prior Spoof Prompt is short as PSP.}
	\label{tabab}
	\centering 
	\begin{tabular}{|p{1.1cm}|p{1.4cm}<{\centering}|p{1.4cm}<{\centering}|p{1.2cm}<{\centering}|p{1.1cm}<{\centering}|
		}
		\hline
		& APCER(\%)$\downarrow$&BPCER(\%)$\downarrow$ &ACER(\%)$\downarrow$ & AUC(\%)$\uparrow$  \\ \hline
		w/o $\ell_{con}$&13.03&0.00&6.51&96.12\\ \hline
		w/o $\ell_{div}$&9.96&11.76&10.86&94.10\\ \hline
		w/o $\ell_{gui}$&14.94&0.00&7.47&96.55\\ \hline
		w/o $\ell_{dc}$&23.75&5.88&14.82&92.54\\ \hline
		w/o USP&6.51&11.76&9.14&95.67\\ \hline
		w/o PSP&4.60&5.88&5.24&97.23\\ \hline
		Ours&9.58&0.00&4.79&98.47\\
		\hline
	\end{tabular}
\end{table}

\subsection{Ablation Study and Hyperparameter Analysis}
\subsubsection{Ablation Study}
In this section, we conduct experiments to evaluate the contribution of each component to our method. Given that the silicone mask is a frequently encountered attack type in real-world face anti-spoofing applications, we perform the analysis using the protocol where the WMCA dataset serves as the source domain, SiW-Mv2 as the target domain, and the silicone mask is left as the attack type.
The results are presented in Table~\ref{tabab}.

In terms of ACER, removing the learnable unknown spoof prompts results in a significant 47.59\% performance degradation, while eliminating prior spoof prompts leads to an 8.59\% decrease. This highlights that the learnable spoof prompts play a crucial role in adapting the vision-language model to the face anti-spoofing task. In contrast, prior knowledge contributes to the overall performance improvement but does not play a decisive impact; rather, it serves as a loose guiding constraint. This finding is further supported by the performance drop observed after removing the prior spoof-guided regularization $\ell_{gui}$. 

The most significant performance decline occurs when the one-class discriminative classification regularization $\ell_{dc}$ and the diverse spoof prompt regularization $\ell_{div}$ are removed, with performance reductions of 67.68\% and 55.89\%, respectively. 
This indicates that explicitly enforcing a discrimination constraint on the similarity between real faces and real prompts is critical for accurate classification of real and spoof faces.
At the same time,  ensuring the diversity of spoof prompts is essential for distinguishing unknown spoof attacks. The spoof prompt contrastive generation regularization $\ell_{con}$ effectively increases the distance between spoof prompts and real face images while reducing the distance between real prompts and real face images, thereby improving the model’s generalization to unknown attacks across different scenarios.

\begin{figure}[!t]
	\centering
	\includegraphics[width=0.4\textwidth]{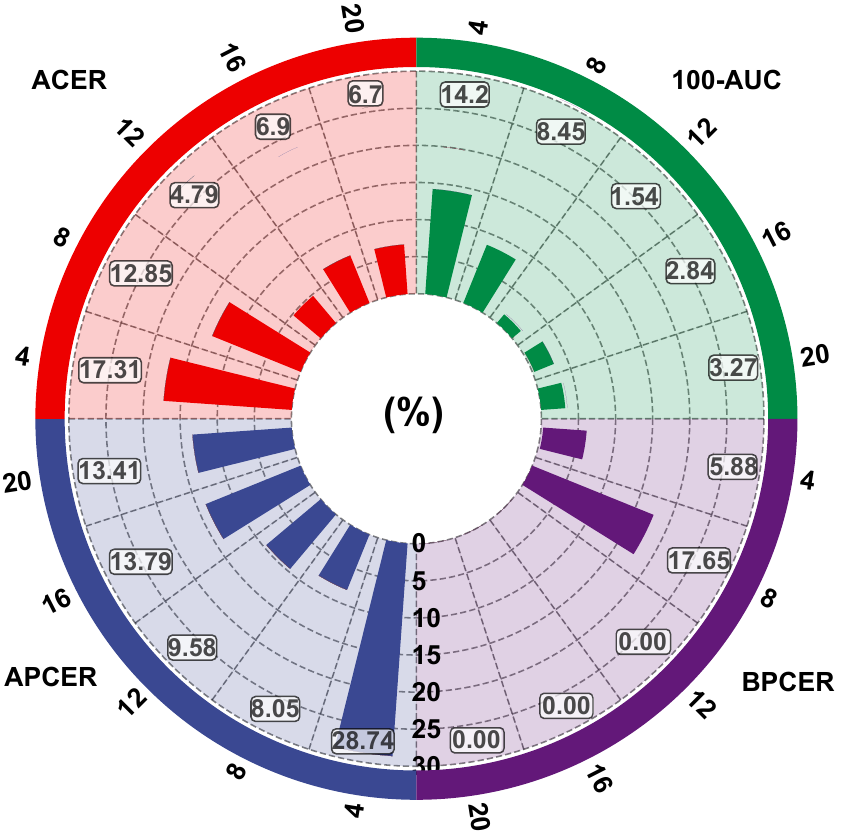}
	\caption{Performance comparison of different number of unknown spoof prompt.
	}
	\label{fig5}
\end{figure}

\begin{figure*}[!t]
	\centering
	\includegraphics[width=1\textwidth]{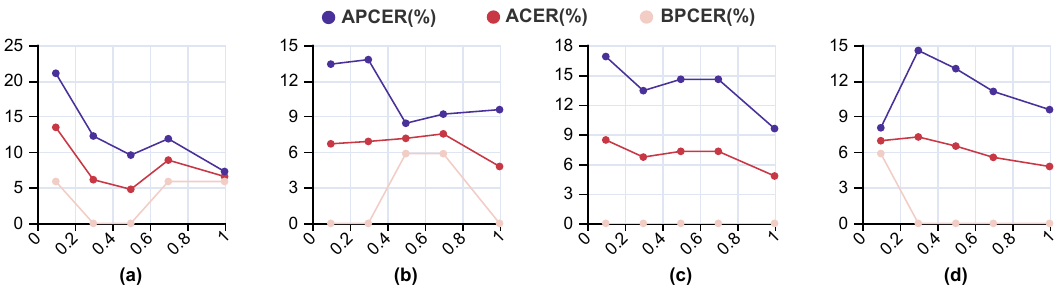}
	\caption{Comparison results of loss balancing parameters with different value in $\in \{0.1,0.3,0.5,0.7,1\}$. (a) $\lambda_1$ for $\ell_{dc}$, (b) $\lambda_2$ for $\ell_{con}$, (c) $\lambda_3$ for $\ell_{div}$, (d) $\lambda_4$ for $\ell_{gui}$.
	}
	\label{fig4}
\end{figure*}

\begin{figure}[!t]
	\centering
	\includegraphics[width=0.5\textwidth]{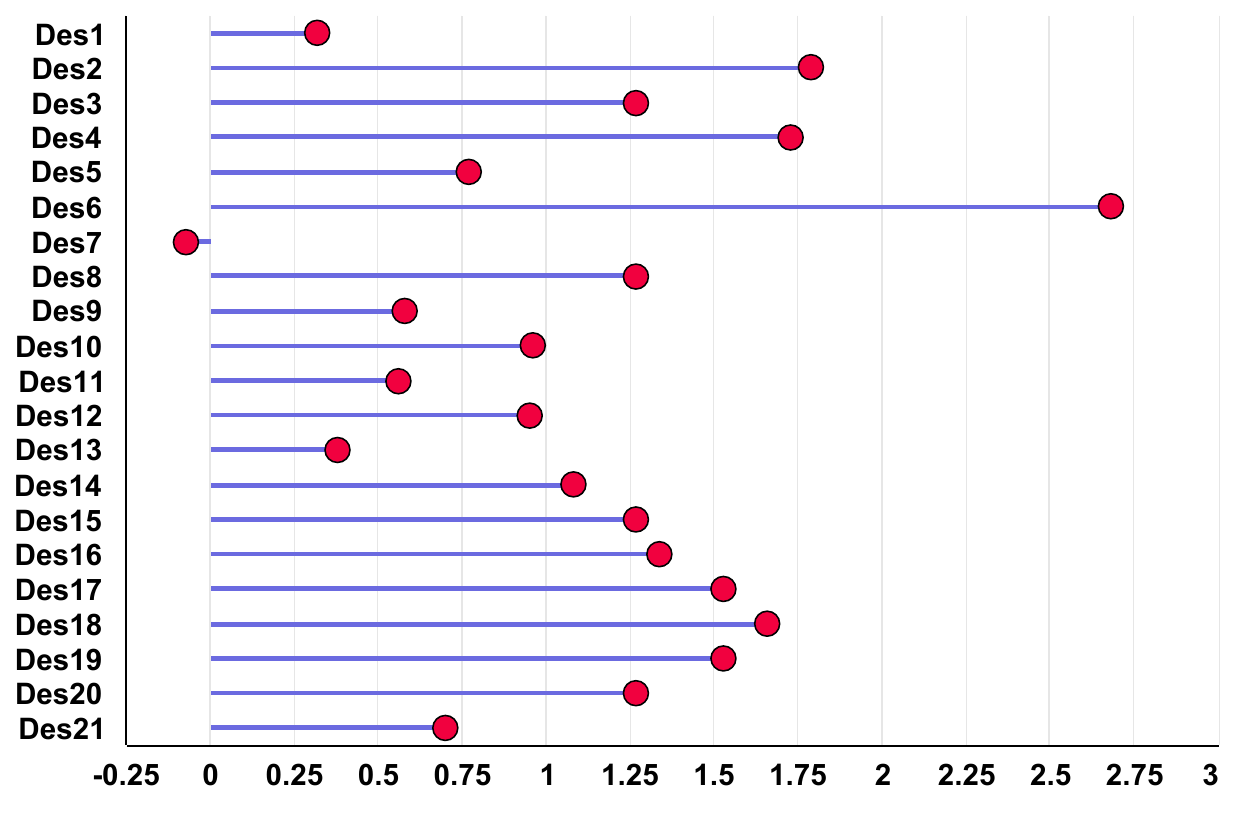}
	\caption{The ACER(\%) performance gap between the model using all prior descriptions and the model without an individual prior description. The model using all prior descriptions achieves an ACER of 4.79\% under the benchmark protocol.
	}
	\label{fig6}
\end{figure}

\subsubsection{Analysis of the Number of Unknown Prompts}
The diversity of unknown spoof prompts is essential for enhancing the model's generalization capability. In this subsection, we analyze the impact of varying the number of unknown spoof prompts, $N^u$, on the generalization performance. The results are shown in \figurename~\ref{fig5}. As $N^u$ increases from 4 to 20, all four evaluation metrics show a gradual improvement, reaching optimal performance when $N^u=12$. Beyond this point, the performance plateaus. This suggests that as $N^u$ increases, the model captures a broader range of potential unknown attack patterns, leading to better generalization. However, an excessively large $N^u$ does not necessarily yield further improvements. Once the prompts encapsulate sufficient contextual information, additional increases in $N^u$ have a negligible impact. Based on the overall model performance, we set $N^u$ as 12 in this paper.

\subsubsection{Effectiveness of an Individual Prior Description}

We compile 21 textual descriptions as prior knowledge to guide the learning of real and unknown spoof prompts, as detailed in Table~\ref{tabDes} of the appendix. To evaluate the contribution of each individual prior description, we conduct experiments using the same protocol as in the ablation study. The results are presented in \figurename~\ref{fig6}. 

Most prior descriptions enhance the generalization ability of the face anti-spoofing model. Notably, the absence of Des6, Des2, and Des4 has the most significant impact on performance, as these descriptions incorporate prior knowledge related to spoofing patterns, including aspects such as material, surface texture, reflection properties, and motion differences. The performance improvement indicates that injecting well-structured prior knowledge effectively facilitates prompt learning. However, a few prior descriptions, such as Des7, result in performance degradation. Des7 introduces depth-related prior knowledge about spoofing patterns, which is not a discriminative feature for the silicone mask. Consequently, its inclusion results in a decline in model performance. This further emphasizes that prior descriptions aligned with the spoofing patterns of unknown attacks contribute to performance enhancement, with prior descriptions playing a key guiding role in the prompt learning process.

\begin{table*}[!tbp]
	\footnotesize
	\caption{Intra-domain leave-one-attack-out evaluation results on the SiW-Mv2 datasets. Red and blue labels represent the best-performing and second-best-performing in terms of ACER.}
	\label{tabintra1}
	\centering 
	\begin{tabular}{|p{1.5cm}<{\centering}|p{1.1cm}<{\centering}|p{0.6cm}<{\centering}|p{0.6cm}<{\centering}|p{0.6cm}<{\centering}|p{0.6cm}<{\centering}|p{0.6cm}<{\centering}|p{0.6cm}<{\centering}|p{0.6cm}<{\centering}|p{0.6cm}<{\centering}|p{0.6cm}<{\centering}|p{0.6cm}<{\centering}|p{0.6cm}<{\centering}|p{0.6cm}<{\centering}|p{0.6cm}<{\centering}|p{0.6cm}<{\centering}|p{0.6cm}<{\centering}|
		}
		\hline	
		\multirow{2}{*}{Method}&\multirow{2}{*}{Metric(\%)}&\multicolumn{5}{c|}{ Mask Attacks}& \multicolumn{4}{c|}{Partial Attacks}&\multicolumn{2}{c|}{2D Attacks}&\multicolumn{3}{c|}{Makeup Attacks}& \multirow{2}{*}{Avg.} \\ \cline{3-16}
		            &&Hal. & Pap.&Tra. &Sil. & Man.&Eye&Fun.&  Mou.&Gla.&Rep.&Print&Cos.&Imp.& Obf.&     \\ \hline		

	\multirow{4}{*}{\parbox{1.5cm}{\centering OC-CNN\\~\citep{oza2018one}}}
		&APCER$\downarrow$&0.00&45.98&88.50&84.29&65.13&27.59&0.38&0.00&0.77&98.85&0.00&0.00&79.31&85.44&41.16         \\ 
		&BPCER$\downarrow$&100.00&41.18&16.67&5.88&30.00&49.12&100.00&100.00&100.00&2.04&100.00&100.00&4.92&31.81&55.83         \\ 
		&ACER$\downarrow$&50.00&43.58&52.59&45.08&47.57&38.35&50.19&50.00&50.38&50.44&50.00&50.00&42.11&58.63&48.49         \\ 
		&AUC$\uparrow$&51.18&59.82&29.33&60.65&55.62&68.99&31.09&61.31&23.65&66.73&70.07&46.40&65.15&31.83&51.56         \\ 
		\hline
		\multirow{4}{*}{\parbox{1.5cm}{\centering AD\\~\citep{baweja2020anomaly}}}
		&APCER$\downarrow$&54.79&43.30&100.00&100.00&39.08&68.97&100.00&100.00&55.55&29.89&100.00&59.00&54.41&61.30&69.02         \\ 
		&BPCER$\downarrow$&52.78&41.18&0.00&0.00&40.00&70.18&0.00&0.00&48.68&28.57&0.00&55.77&59.02&59.09&32.52         \\ 
		&ACER$\downarrow$&53.78&42.24&50.00&50.00&39.54&69.57&50.00&50.00&52.12&29.23&50.00&57.39&56.71&60.20&50.77         \\ 
		&AUC$\uparrow$&45.96&58.03&46.11&59.22&63.80&25.66&39.86&59.37&44.49&76.10&57.55&37.95&45.07&32.10&49.38         \\
		\hline
		\multirow{4}{*}{\parbox{1.5cm}{\centering Hyp-OC\\~\citep{narayan2024hyp}}}
		&APCER$\downarrow$&31.80&11.11&38.31&10.72&24.90&17.62&41.76&9.58&26.82&16.86&22.22&24.52&15.33&24.52& 22.58        \\ 
		&BPCER$\downarrow$&33.33&17.65&36.67&17.64&10.00&17.54&34.08&17.24&44.74&17.35&21.48&25.00&19.67&27.27&24.26         \\ 
		&ACER$\downarrow$&\textcolor{blue}{32.57}&\textcolor{blue}{14.38}&37.49&\textcolor{blue}{14.18}&\textcolor{blue}{17.45}&\textcolor{blue}{17.58}&37.92&\textcolor{blue}{13.41}&35.78&17.10&\textcolor{blue}{21.85}&\textcolor{blue}{24.76}&\textcolor{blue}{17.50}&\textcolor{blue}{25.90}&\textcolor{blue}{23.42} \\ 
		&AUC$\uparrow$&71.10&92.81&62.91&88.30&89.69&87.83&64.68&91.74&71.09&89.18&87.01&81.45&89.99&79.35& 81.94\\
		
		\hline
		\multirow{4}{*}{\parbox{1.5cm}{\centering OC-SCMNet\\~\citep{huang2024one}}}
		&APCER$\downarrow$&19.44&23.53&30.00&11.76&37.50&17.54&16.76&3.45&56.58&16.33&58.51&34.62&14.75&18.18&         25.64\\ 
		&BPCER$\downarrow$&47.51&25.28&30.27&40.23&28.74&23.75&43.67&45.21&12.64&16.86&12.26&27.97&38.70&38.69&         30.84\\	 
		
		&ACER$\downarrow$&33.48&24.41&\textcolor{blue}{30.13}&26.00&33.12&20.65&\textcolor{red}{30.22}&24.33&\textcolor{blue}{34.61}&\textcolor{blue}{16.59}&35.39&31.29&26.73&28.44&         28.24\\	 
		&AUC$\uparrow$&59.05&71.42&71.55&73.68&69.89&82.13&66.08&75.15&65.43&86.88&68.14&67.85&76.35&69.84&         71.67\\	
		\hline
		\multirow{5}{*}{Ours}&APCER$\downarrow$&34.10&14.18&18.39& 0.76&3.45&   3.83&41.38&6.13&21.83&3.45&3.83&3.45&0.00&40.99&13.98\\ 
		&BPCER$\downarrow$&15.28&0.00&5.00& 0.00&5.00&0.00&29.05&0.00&26.31&10.20&4.44&5.00&0.00&4.54&7.49\\ 
		&ACER$\downarrow$&\textcolor{red}{24.69}&\textcolor{red}{7.09}&\textcolor{red}{11.70}&\textcolor{red}{0.38}&\textcolor{red}{4.22}&\textcolor{red}{1.92}&\textcolor{blue}{35.21}&\textcolor{red}{3.06}&\textcolor{red}{24.08}&\textcolor{red}{6.83}&\textcolor{red}{4.14}&\textcolor{red}{23.77}&\textcolor{red}{0.00}&\textcolor{red}{22.78}&\textcolor{red}{12.13}\\	
		&AUC$\uparrow$&80.81&96.71&94.78&99.95&98.91&98.14&68.27&97.08&82.10&96.94&99.08&78.30&100.00&79.95&90.79\\			
		\hline
	\end{tabular}
\end{table*}

\begin{table*}[!tbp]
	\footnotesize
	\caption{Intra-domain leave-one-attack-out evaluation results on the WMCA datasets. Red and blue labels represent the best-performing and second-best-performing in terms of ACER.}
	\label{tabintra2}
	\centering 
    \begin{tabular}{|p{4.25cm}|p{1.15cm}<{\centering}|p{1.15cm}<{\centering}|p{1.15cm}<{\centering}|p{1.15cm}<{\centering}|p{1.15cm}<{\centering}|p{1.15cm}<{\centering}|p{1.15cm}<{\centering}|p{1.15cm}<{\centering}|p{1.2cm}<{\centering}|}
		\hline
		Method& Metric (\%)  & Fake Head  & Paper Mask  & Rigid Mask & Flexible Mask  & Glasses & Replay& Print&Average\\ \hline
		
		\multirow{4}{*}{OC-CNN~\citep{oza2018one}}
		&APCER$\downarrow$&2.61&13.91&0.00&86.09&3.48&34.78&96.91&33.97         \\
		&BPCER$\downarrow$&98.36&90.14&100.00&24.80&94.39&72.32&6.08&69.44         \\
		&ACER$\downarrow$&50.48&52.03&50.00&55.44&48.93&53.55&51.50&51.70         \\
		&AUC$\uparrow$&41.79&40.20&53.19&40.28&46.70&49.62&54.29&46.58         \\
		\hline
		\multirow{4}{*}{AD~\citep{baweja2020anomaly}}
		&APCER$\downarrow$&100.00&61.74&60.87&100.00&59.13&100.00&73.04&79.25         \\
		&BPCER$\downarrow$&0.00&60.56&57.14&0.00&62.62&0.00&72.20& 36.07        \\
		&ACER$\downarrow$&50.00&61.15&59.00&50.00&60.87&50.00&72.62&57.66         \\
		&AUC$\uparrow$&53.23&36.15&40.46&69.49&32.45&59.42&21.54& 44.68        \\
		\hline
		\multirow{4}{*}{Hyp-OC~\citep{narayan2024hyp}}
		&APCER$\downarrow$&26.09&26.09&7.83&51.30&28.70&14.78&4.35& 22.74        \\
		&BPCER$\downarrow$&27.87&28.17&40.00&17.41&28.97&14.19&10.42&23.86         \\
		&ACER$\downarrow$&\textcolor{blue}{26.97}&\textcolor{blue}{27.13}&\textcolor{blue}{23.91}&34.36&\textcolor{blue}{28.83}&14.48&\textcolor{blue}{7.39}&\textcolor{blue}{23.30}         \\
		&AUC$\uparrow$&83.94&81.74&86.36&71.47&76.63&88.83&95.95& 83.56        \\
		\hline
		\multirow{4}{*}{OC-SCMNet~\citep{huang2024one}}
		&APCER$\downarrow$&50.82&30.99&57.14&56.20&26.17&11.76&32.05& 37.88        \\				
		&BPCER$\downarrow$&27.83&46.09&8.70&11.30&35.65&16.52&3.48& 21.37        \\				
		&ACER$\downarrow$&39.32&38.54&32.92&\textcolor{blue}{33.75}&30.91&\textcolor{blue}{14.14}&17.76&29.62         \\				
		&AUC$\uparrow$&53.19&58.80&70.61&62.78&69.11&90.85&82.27&69.66         \\	
		\hline
		\multirow{4}{*}{Ours}
		&APCER$\downarrow$&1.74&0.00&12.07&7.83&7.83&10.43&1.74&5.95\\
		&BPCER$\downarrow$&0.82&0.00&1.90&5.54&7.48&0.00&0.77&2.36\\	 
        &ACER$\downarrow$&\textcolor{red}{1.28}&\textcolor{red}{0.00}&\textcolor{red}{7.04}&\textcolor{red}{6.68}&\textcolor{red}{7.65}&\textcolor{red}{5.22}&\textcolor{red}{1.26}&\textcolor{red}{4.16}\\		
		&AUC$\uparrow$&99.86&100.00&97.45&98.33&97.55&98.44&99.65&98.75\\			
		\hline
	\end{tabular}
\end{table*}

\subsubsection{Analysis of Loss Balancing Parameters}
In this subsection, we perform a hyper-parameter analysis of $\lambda_1$, $\lambda_2$, $\lambda_3$, and $\lambda_4$ using the same protocol as in the ablation study. The results are shown in \figurename~\ref{fig4}. For $\lambda_1$, as its value increases from 0 to 1, performance is initially improved, reaching the optimal performance at $\lambda_1 = 0.5$, after which it gradually decreases. Therefore, we set $\lambda_1$ to 0.5. Similarly, for $\lambda_2$, $\lambda_3$, and $\lambda_4$, performance is consistently improved as their values increase from 0 to 1, with the best results achieved when they are set to 1. Hence, we set $\lambda_2$, $\lambda_3$, and $\lambda_4$ to 1.

\subsubsection{Analysis of Covariate Shift and Semantic Shift}

There exist significant covariate shift and semantic shift between the unknown target domain and the source domain. To evaluate the impact of these shifts on the generalization performance of face anti-spoofing models, we conduct intra-domain leave-one-attack-out experiments using the SiW-Mv2 and WMCA datasets. For the SiW-Mv2 dataset, we follow the official definition of Protocol II. For the WMCA dataset, we adhere to the official 7:3 ratio for randomly splitting the training and testing sets. The experimental results are presented in Table~\ref{tabintra1} and Table~\ref{tabintra2}.

The intra-domain leave-one-attack-out protocol mitigates the impact of covariate shift and primarily evaluates the model’s generalization ability to semantic shift. The results show that our method significantly improves the generalization capability for detecting diverse unknown attack types compared to existing state-of-the-art methods, regardless of whether it is tested on the SiW-Mv2 or WMCA dataset. In terms of ACER, our approach improves the average performance across 14 attack types in the SiW-Mv2 dataset by 48.21\% and across 7 attack types in the WMCA dataset by 82.15\% compared to previous state-of-the-art methods.  
These improvements demonstrate that our method effectively learns spoof prompts for diverse attack types, enabling better handling of semantic shift.

\begin{figure}[!t]
	\centering
	\includegraphics[width=0.49\textwidth]{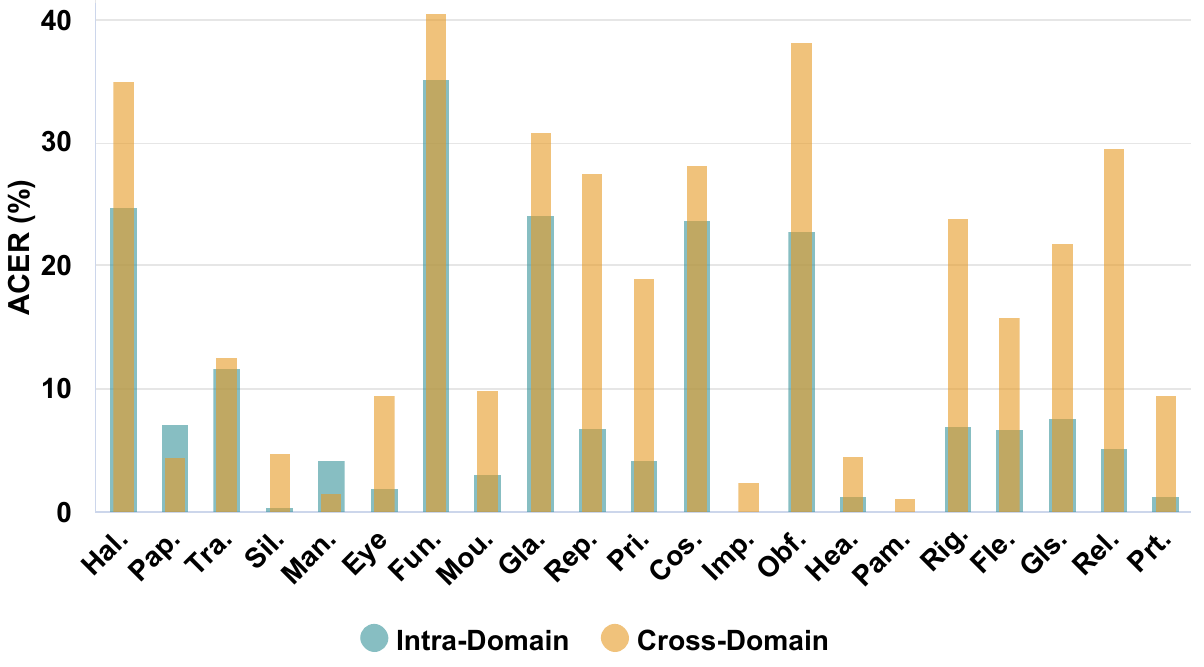}
	\caption{The intra-domain and cross-domain leave-one-attack-out ACER(\%) performance comparison of our method on the SiW-Mv2 and WMCA datasets.
	}
	\label{fig10}
\end{figure}

We further analyze the impact of covariate shift on model performance by synthesizing results from Table~\ref{tabintra1}, Table~\ref{tabintra2}, Tables~\ref{tabcross1}, and Tables~\ref{tabcross2}. To facilitate comparison, we visualize the ACER values of our method under intra-domain and cross-domain leave-one-attack-out protocols in \figurename~\ref{fig10}. For most unknown attack types, covariate shift results in a decline in model performance. A detailed analysis of the APCER and BPCER distributions reveals that covariate shift has a more pronounced impact on BPCER. External factors such as recording background, lighting conditions, and recording devices introduce spoof-irrelevant appearance variations between real face images in the source and target domains. This covariate shift increases the likelihood of misclassification for real faces in the target domain, leading to a higher BPCER. This, in turn, reduces the overall detection performance in terms of ACER. However, compared to existing state-of-the-art methods, the performance improvement of our approach on cross-domain leave-one-attack-out protocols demonstrates its superior ability to handle the challenges posed by covariate shift. This also suggests that our method has learned effective textual prompts and successfully transferred the general knowledge of the vision-language model to mitigate the impact of covariate shift.

\section{Conclusion}

In this paper, we propose a novel unknown spoof prompt learning method for generalized face anti-spoofing. This method learns fine-grained and diverse textual prompts using only real face images from a single source domain, adapting vision-language models to generalize across cross-scenario target domains where both covariate shift and semantic shift coexist.
Extensive experiments on nine benchmark datasets demonstrate that our method significantly enhances the generalization ability of face anti-spoofing models against diverse unknown attack types in unseen scenarios, all while maintaining a low training data cost. %
Despite these advancements, our method still faces challenges in distinguishing highly realistic makeup attacks. In future work, we aim to learn reflection properties and material characteristics into spoof prompts, further enhancing the generalization ability of face anti-spoofing models against various attack types.
\section*{Data availability}
The data that support the findings of this study are available from the authors upon reasonable request.

\begin{acknowledgements}
This work was supported in part by the National Key Research and Development Program of China (Grant No. 2022YFC3310400), in part by the Natural Science Foundation of China (Grant Nos. U23B2054, 62102419, 62276263 and 62406133), in part by the Youth Innovation Promotion Association CAS (Grant No. Y2023143), in part by the Natural Science Foundation of Hunan Province (Grant No. 2024JJ6389) and in part by the Hengyang Science and Technology Plan Project (Grant No. 202330046190).
			
\end{acknowledgements}

{
\small
\bibliographystyle{spbasic}      
\bibliography{ref2}

\begin{thebibliography}{71}
\providecommand{\natexlab}[1]{#1}
\providecommand{\url}[1]{{#1}}
\providecommand{\urlprefix}{URL }
\expandafter\ifx\csname urlstyle\endcsname\relax
  \providecommand{\doi}[1]{DOI~\discretionary{}{}{}#1}\else
  \providecommand{\doi}{DOI~\discretionary{}{}{}\begingroup
  \urlstyle{rm}\Url}\fi
\providecommand{\eprint}[2][]{\url{#2}}

\bibitem[{Abduh and Ivrissimtzis(2020)}]{abduh2020use}
Abduh L, Ivrissimtzis I (2020) Use of in-the-wild images for anomaly detection
  in face anti-spoofing. arXiv preprint arXiv:200610626

\bibitem[{Achiam et~al.(2023)Achiam, Adler, Agarwal, Ahmad, Akkaya, Aleman,
  Almeida, Altenschmidt, Altman, Anadkat et~al.}]{achiam2023gpt}
Achiam J, Adler S, Agarwal S, Ahmad L, Akkaya I, Aleman FL, Almeida D,
  Altenschmidt J, Altman S, Anadkat S, et~al. (2023) Gpt-4 technical report.
  arXiv preprint arXiv:230308774

\bibitem[{Arashloo(2020)}]{arashloo2020unseen}
Arashloo SR (2020) Unseen face presentation attack detection using sparse
  multiple kernel fisher null-space. IEEE Transactions on Circuits and Systems
  for Video Technology 31(10):4084--4095

\bibitem[{Arashloo(2023)}]{arashloo2023unknown}
Arashloo SR (2023) Unknown face presentation attack detection via localized
  learning of multiple kernels. IEEE Transactions on Information Forensics and
  Security 18:1421--1432

\bibitem[{Bahng et~al.(2022)Bahng, Jahanian, Sankaranarayanan, and
  Isola}]{bahng2022exploring}
Bahng H, Jahanian A, Sankaranarayanan S, Isola P (2022) Exploring visual
  prompts for adapting large-scale models. arXiv preprint arXiv:220317274

\bibitem[{Baweja et~al.(2020)Baweja, Oza, Perera, and
  Patel}]{baweja2020anomaly}
Baweja Y, Oza P, Perera P, Patel VM (2020) Anomaly detection-based unknown face
  presentation attack detection. In: IEEE International Joint Conference on
  Biometrics, pp 1--9

\bibitem[{Boulkenafet et~al.(2015)Boulkenafet, Komulainen, and
  Hadid}]{mmboulkenafet2015face}
Boulkenafet Z, Komulainen J, Hadid A (2015) Face anti-spoofing based on color
  texture analysis. In: IEEE International Conference on Image Processing, pp
  2636--2640

\bibitem[{Boulkenafet et~al.(2017)Boulkenafet, Komulainen, Li, Feng, and
  Hadid}]{boulkenafet2017oulu}
Boulkenafet Z, Komulainen J, Li L, Feng X, Hadid A (2017) Oulu-npu: A mobile
  face presentation attack database with real-world variations. In: IEEE
  International Conference on Automatic Face and Gesture Recognition, pp
  612--618

\bibitem[{Cai et~al.(2022)Cai, Li, Wan, Li, Hu, and Kot}]{cai2022learning}
Cai R, Li Z, Wan R, Li H, Hu Y, Kot AC (2022) Learning meta pattern for face
  anti-spoofing. IEEE Transactions on Information Forensics and Security
  17:1201--1213

\bibitem[{Cai et~al.(2024{\natexlab{a}})Cai, Soh, Yu, Li, Yang, and
  Kot}]{cai2024towards}
Cai R, Soh C, Yu Z, Li H, Yang W, Kot AC (2024{\natexlab{a}}) Towards
  data-centric face anti-spoofing: Improving cross-domain generalization via
  physics-based data synthesis. International Journal of Computer Vision pp
  1--22

\bibitem[{Cai et~al.(2024{\natexlab{b}})Cai, Yu, Kong, Li, Chen, Hu, and
  Kot}]{cai2024s}
Cai R, Yu Z, Kong C, Li H, Chen C, Hu Y, Kot AC (2024{\natexlab{b}}) S-adapter:
  Generalizing vision transformer for face anti-spoofing with statistical
  tokens. IEEE Transactions on Information Forensics and Security

\bibitem[{Chingovska et~al.(2012)Chingovska, Anjos, and
  Marcel}]{chingovska2012effectiveness}
Chingovska I, Anjos A, Marcel S (2012) On the effectiveness of local binary
  patterns in face anti-spoofing. In: International Conference of Biometrics
  Special Interest Group, pp 1--7

\bibitem[{Du et~al.(2022)Du, Li, Zuo, Zhu, and Lu}]{du2022energy}
Du Z, Li J, Zuo L, Zhu L, Lu K (2022) Energy-based domain generalization for
  face anti-spoofing. In: ACM International Conference on Multimedia, pp
  1749--1757

\bibitem[{Erdogmus and Marcel(2014)}]{erdogmus2014spoofing2}
Erdogmus N, Marcel S (2014) Spoofing face recognition with 3d masks. IEEE
  Transactions on Information Forensics and Security 9(7):1084--1097

\bibitem[{Fang et~al.(2024)Fang, Liu, Jiang, Lu, Zhao, and Wan}]{fang2024vl}
Fang H, Liu A, Jiang N, Lu Q, Zhao G, Wan J (2024) Vl-fas: Domain
  generalization via vision-language model for face anti-spoofing. In: IEEE
  International Conference on Acoustics, Speech and Signal Processing, pp
  4770--4774

\bibitem[{Fatemifar et~al.(2019)Fatemifar, Awais, Arashloo, and
  Kittler}]{fatemifar2019combining}
Fatemifar S, Awais M, Arashloo SR, Kittler J (2019) Combining multiple
  one-class classifiers for anomaly based face spoofing attack detection. In:
  International Conference on Biometrics, pp 1--7

\bibitem[{Fatemifar et~al.(2020)Fatemifar, Awais, Akbari, and
  Kittler}]{fatemifar2020stacking}
Fatemifar S, Awais M, Akbari A, Kittler J (2020) A stacking ensemble for
  anomaly based client-specific face spoofing detection. In: IEEE International
  Conference on Image Processing, pp 1371--1375

\bibitem[{Ge et~al.(2024)Ge, Liu, Yu, Shi, Qi, Li, and
  K{\"a}lvi{\"a}inen}]{ge2024difffas}
Ge X, Liu X, Yu Z, Shi J, Qi C, Li J, K{\"a}lvi{\"a}inen H (2024) Difffas: face
  anti-spoofing via generative diffusion models. In: European Conference on
  Computer Vision, pp 144--161

\bibitem[{George et~al.(2019)George, Mostaani, Geissenbuhler, Nikisins, Anjos,
  and Marcel}]{george2019biometric}
George A, Mostaani Z, Geissenbuhler D, Nikisins O, Anjos A, Marcel S (2019)
  Biometric face presentation attack detection with multi-channel convolutional
  neural network. IEEE Transactions on Information Forensics and Security
  15:42--55

\bibitem[{Guo et~al.(2024)Guo, Liu, Luo, Hu, Zou, Zhang, Liu, and
  Zhao}]{guo2024style}
Guo J, Liu H, Luo Y, Hu X, Zou H, Zhang Y, Liu H, Zhao B (2024)
  Style-conditional prompt token learning for generalizable face anti-spoofing.
  In: ACM International Conference on Multimedia, pp 994--1003

\bibitem[{Guo et~al.(2022)Guo, Liu, Jain, and Liu}]{guo2022multi}
Guo X, Liu Y, Jain A, Liu X (2022) Multi-domain learning for updating face
  anti-spoofing models. In: European Conference on Computer Vision, pp 230--249

\bibitem[{Gwon and Kim(2024)}]{gwon2024one}
Gwon MG, Kim W (2024) One-class learning for face anti-spoofing via
  pseudo-negative sampling. Multimedia Tools and Applications
  83(18):54791--54813

\bibitem[{Hu et~al.(2024{\natexlab{a}})Hu, Zhang, Yao, Ding, and
  Ma}]{hu2024rethinking}
Hu C, Zhang KY, Yao T, Ding S, Ma L (2024{\natexlab{a}}) Rethinking
  generalizable face anti-spoofing via hierarchical prototype-guided
  distribution refinement in hyperbolic space. In: IEEE Conference on Computer
  Vision and Pattern Recognition, pp 1032--1041

\bibitem[{Hu et~al.(2024{\natexlab{b}})Hu, Liu, Yuan, Fu, Luo, Zhang, Zou, Gan,
  and Zhang}]{hu2024fine}
Hu X, Liu H, Yuan H, Fu Z, Luo Y, Zhang N, Zou H, Gan J, Zhang Y
  (2024{\natexlab{b}}) Fine-grained prompt learning for face anti-spoofing. In:
  ACM International Conference on Multimedia, pp 7619--7628

\bibitem[{Huang et~al.(2024)Huang, Chiang, Chen, Chong, Liu, and
  Hsu}]{huang2024one}
Huang PK, Chiang CH, Chen TH, Chong JX, Liu TL, Hsu CT (2024) One-class face
  anti-spoofing via spoof cue map-guided feature learning. In: IEEE Conference
  on Computer Vision and Pattern Recognition, pp 277--286

\bibitem[{Huang et~al.(2021)Huang, Xia, and Shen}]{huang2021one}
Huang X, Xia J, Shen L (2021) One-class face anti-spoofing based on attention
  auto-encoder. In: Chinese Conference Biometric Recognition, pp 365--373

\bibitem[{Jia et~al.(2020)Jia, Zhang, Shan, and Chen}]{jia2020single}
Jia Y, Zhang J, Shan S, Chen X (2020) Single-side domain generalization for
  face anti-spoofing. In: IEEE Conference on Computer Vision and Pattern
  Recognition, pp 8484--8493

\bibitem[{Jia et~al.(2021)Jia, Zhang, and Shan}]{jia2021dual}
Jia Y, Zhang J, Shan S (2021) Dual-branch meta-learning network with
  distribution alignment for face anti-spoofing. IEEE Transactions on
  Information Forensics and Security 17:138--151

\bibitem[{Jiang et~al.(2023)Jiang, Li, Liu, Zhou, and
  Sun}]{jiang2023adversarial}
Jiang F, Li Q, Liu P, Zhou XD, Sun Z (2023) Adversarial learning
  domain-invariant conditional features for robust face anti-spoofing.
  International Journal of Computer Vision 131:1680--1703

\bibitem[{Jiang et~al.(2024)Jiang, Liu, Si, Meng, and Li}]{jiang2024cross}
Jiang F, Liu Y, Si H, Meng J, Li Q (2024) Cross-scenario unknown-aware face
  anti-spoofing with evidential semantic consistency learning. IEEE
  Transactions on Information Forensics and Security 19:3093--3108

\bibitem[{Khattak et~al.(2023)Khattak, Rasheed, Maaz, Khan, and
  Khan}]{khattak2023maple}
Khattak MU, Rasheed H, Maaz M, Khan S, Khan FS (2023) Maple: Multi-modal prompt
  learning. In: IEEE Conference on Computer Vision and Pattern Recognition, pp
  19113--19122

\bibitem[{Kong et~al.(2024)Kong, Zhang, Wang, Zhang, Li, Tang, and
  Luo}]{kong2024dual}
Kong Z, Zhang W, Wang T, Zhang K, Li Y, Tang X, Luo W (2024) Dual teacher
  knowledge distillation with domain alignment for face anti-spoofing. IEEE
  Transactions on Circuits and Systems for Video Technology

\bibitem[{Le and Woo(2024)}]{le2024gradient}
Le BM, Woo SS (2024) Gradient alignment for cross-domain face anti-spoofing.
  In: IEEE Conference on Computer Vision and Pattern Recognition, pp 188--199

\bibitem[{Li et~al.(2018)Li, He, Wang, Rocha, Jiang, and Kot}]{li2018learning}
Li H, He P, Wang S, Rocha A, Jiang X, Kot AC (2018) Learning generalized deep
  feature representation for face anti-spoofing. IEEE Transactions on
  Information Forensics and Security 13(10):2639--2652

\bibitem[{Liu(2024)}]{liu2024moeit}
Liu A (2024) Ca-moeit: Generalizable face anti-spoofing via dual
  cross-attention and semi-fixed mixture-of-expert. International Journal of
  Computer Vision 132(11):5439--5452

\bibitem[{Liu et~al.(2024{\natexlab{a}})Liu, Xue, Gan, Wan, Liang, Deng,
  Escalera, and Lei}]{liu2024cfpl}
Liu A, Xue S, Gan J, Wan J, Liang Y, Deng J, Escalera S, Lei Z
  (2024{\natexlab{a}}) Cfpl-fas: Class free prompt learning for generalizable
  face anti-spoofing. In: Conference on Computer Vision and Pattern
  Recognition, pp 222--232

\bibitem[{Liu et~al.(2016)Liu, Yuen, Zhang, and Zhao}]{liu20163dr}
Liu S, Yuen PC, Zhang S, Zhao G (2016) 3d mask face anti-spoofing with remote
  photoplethysmography. In: European Conference on Computer Vision, pp 85--100

\bibitem[{Liu et~al.(2024{\natexlab{b}})Liu, Wang, and Yuen}]{liu2024bottom}
Liu SQ, Wang Q, Yuen PC (2024{\natexlab{b}}) Bottom-up domain prompt tuning for
  generalized face anti-spoofing. In: European Conference on Computer Vision,
  pp 170--187

\bibitem[{Liu et~al.(2024{\natexlab{c}})Liu, Li, Xu, Guo, Zou, and
  Wu}]{liu2024quality}
Liu Y, Li Z, Xu Y, Guo Z, Zou Z, Wu L (2024{\natexlab{c}}) Quality-invariant
  domain generalization for face anti-spoofing. International Journal of
  Computer Vision 132(11):5239--5254

\bibitem[{Liu et~al.(2025)Liu, Li, and Wu}]{liu2025dual}
Liu Y, Li Z, Wu L (2025) Dual consistency regularization for generalized face
  anti-spoofing. IEEE Transactions on Information Forensics and Security

\bibitem[{Long et~al.(2024)Long, Zhang, and Shan}]{long2024confidence}
Long X, Zhang J, Shan S (2024) Confidence aware learning for reliable face
  anti-spoofing. arXiv preprint arXiv:241101263

\bibitem[{Ma et~al.(2024)Ma, Qian, Li, and Yang}]{ma2024dual}
Ma Y, Qian J, Li J, Yang J (2024) Dual feature disentanglement for face
  anti-spoofing. Pattern Recognition 155:110656

\bibitem[{Mu et~al.(2023)Mu, Bai, He, Ye, Liang, Yang, Zhuang, and
  Hu}]{mu2023teg}
Mu L, Bai J, He X, Ye J, Liang X, Yang Y, Zhuang J, Hu H (2023) Teg-dg:
  Textually guided domain generalization for face anti-spoofing. arXiv preprint
  arXiv:231118420

\bibitem[{Muhammad et~al.(2023)Muhammad, Beddiar, and
  Oussalah}]{muhammad2023domain}
Muhammad U, Beddiar DR, Oussalah M (2023) Domain generalization via ensemble
  stacking for face presentation attack detection. arXiv preprint
  arXiv:230102145

\bibitem[{Narayan and Patel(2024)}]{narayan2024hyp}
Narayan K, Patel VM (2024) Hyp-oc: Hyperbolic one class classification for face
  anti-spoofing. arXiv preprint arXiv:240414406

\bibitem[{Nikisins et~al.(2018)Nikisins, Mohammadi, Anjos, and
  Marcel}]{nikisins2018effectiveness}
Nikisins O, Mohammadi A, Anjos A, Marcel S (2018) On effectiveness of anomaly
  detection approaches against unseen presentation attacks in face
  anti-spoofing. In: International Conference on Biometrics, pp 75--81

\bibitem[{Oza and Patel(2018)}]{oza2018one}
Oza P, Patel VM (2018) One-class convolutional neural network. IEEE Signal
  Processing Letters 26(2):277--281

\bibitem[{Park and Kim(2023)}]{park2023visual}
Park SM, Kim YG (2023) Visual language integration: A survey and open
  challenges. Computer Science Review 48:100548

\bibitem[{Shao et~al.(2019)Shao, Lan, Li, and Yuen}]{shao2019multi}
Shao R, Lan X, Li J, Yuen PC (2019) Multi-adversarial discriminative deep
  domain generalization for face presentation attack detection. In: IEEE
  Conference on Computer Vision and Pattern Recognition, pp 10023--10031

\bibitem[{Shao et~al.(2020)Shao, Lan, and Yuen}]{shao2020regularized}
Shao R, Lan X, Yuen PC (2020) Regularized fine-grained meta face anti-spoofing.
  In: Association for the Advancement of Artificial Intelligence, pp
  11974--11981

\bibitem[{Srivatsan et~al.(2023)Srivatsan, Naseer, and
  Nandakumar}]{srivatsan2023flip}
Srivatsan K, Naseer M, Nandakumar K (2023) Flip: Cross-domain face
  anti-spoofing with language guidance. In: International Conference on
  Computer Vision, pp 19685--19696

\bibitem[{Sun et~al.(2023)Sun, Liu, Liu, Li, and Chu}]{sun2023rethinking}
Sun Y, Liu Y, Liu X, Li Y, Chu WS (2023) Rethinking domain generalization for
  face anti-spoofing: separability and alignment. arXiv preprint
  arXiv:230313662

\bibitem[{Wang et~al.(2023)Wang, Lan, Liu, Ouyang, Qin, Lu, Chen, Zeng, and
  Yu}]{wang2022generalizing}
Wang J, Lan C, Liu C, Ouyang Y, Qin T, Lu W, Chen Y, Zeng W, Yu P (2023)
  Generalizing to unseen domains: A survey on domain generalization. IEEE
  Transactions on Knowledge and Data Engineering 35(08):8052--8072

\bibitem[{Wang et~al.(2024{\natexlab{a}})Wang, Zhang, Yue, Liang, Huang, Zhang,
  Han, Ding, and Wang}]{wang2024csdg}
Wang K, Zhang G, Yue H, Liang Y, Huang M, Zhang G, Han J, Ding E, Wang J
  (2024{\natexlab{a}}) Csdg-fas: Closed-space domain generalization for face
  anti-spoofing. International Journal of Computer Vision 132(11):4866--4879

\bibitem[{Wang et~al.(2024{\natexlab{b}})Wang, Zhang, Yao, Zhou, Ding, Dai, and
  Ji}]{wang2024tf}
Wang X, Zhang KY, Yao T, Zhou Q, Ding S, Dai P, Ji R (2024{\natexlab{b}})
  Tf-fas: twofold-element fine-grained semantic guidance for generalizable face
  anti-spoofing. In: European Conference on Computer Vision, pp 148--168

\bibitem[{Wang et~al.(2022)Wang, Wang, Yu, Deng, Li, Gao, and
  Wang}]{wang2022domain}
Wang Z, Wang Z, Yu Z, Deng W, Li J, Gao T, Wang Z (2022) Domain generalization
  via shuffled style assembly for face anti-spoofing. In: Proceedings of IEEE
  Conference on Computer Vision and Pattern Recognition, pp 4123--4133

\bibitem[{Wen et~al.(2015)Wen, Han, and Jain}]{wen2015face}
Wen D, Han H, Jain AK (2015) Face spoof detection with image distortion
  analysis. IEEE Transactions on Information Forensics and Security
  10(4):746--761

\bibitem[{Yang et~al.(2024)Yang, Yu, Ni, He, and Li}]{yang2024generalized}
Yang J, Yu Z, Ni X, He J, Li H (2024) Generalized face anti-spoofing via finer
  domain partition and disentangling liveness-irrelevant factors. In: European
  Conference on Artificial Intelligence, pp 274--281

\bibitem[{Yi et~al.(2014)Yi, Lei, Zhang, and Li}]{yi2014face}
Yi D, Lei Z, Zhang Z, Li SZ (2014) Face anti-spoofing: Multi-spectral approach.
  In: Handbook of Biometric Anti-Spoofing, pp 83--102

\bibitem[{Yu et~al.(2020)Yu, Wan, Qin, Li, Li, and Zhao}]{yu2020fas}
Yu Z, Wan J, Qin Y, Li X, Li SZ, Zhao G (2020) Nas-fas: Static-dynamic central
  difference network search for face anti-spoofing. IEEE Transactions on
  Pattern Analysis and Machine Intelligence 43(9):3005--3023

\bibitem[{Zang et~al.(2022)Zang, Li, Zhou, Huang, and Loy}]{zang2022unified}
Zang Y, Li W, Zhou K, Huang C, Loy CC (2022) Unified vision and language prompt
  learning. arXiv preprint arXiv:221007225

\bibitem[{Zhang et~al.(2024{\natexlab{a}})Zhang, Du, Li, Zhu, and
  Shen}]{zhang2024domain}
Zhang D, Du Z, Li J, Zhu L, Shen HT (2024{\natexlab{a}}) Domain-adaptive
  energy-based models for generalizable face anti-spoofing. IEEE Transactions
  on Multimedia

\bibitem[{Zhang et~al.(2025)Zhang, Wang, Yue, Liu, Zhang, Yao, Ding, and
  Wang}]{zhang2025interpretable}
Zhang G, Wang K, Yue H, Liu A, Zhang G, Yao K, Ding E, Wang J (2025)
  Interpretable face anti-spoofing: Enhancing generalization with multimodal
  large language models. arXiv preprint arXiv:250101720

\bibitem[{Zhang et~al.(2024{\natexlab{b}})Zhang, Huang, Jin, and
  Lu}]{zhang2024vision}
Zhang J, Huang J, Jin S, Lu S (2024{\natexlab{b}}) Vision-language models for
  vision tasks: A survey. IEEE Transactions on Pattern Analysis and Machine
  Intelligence

\bibitem[{Zhang et~al.(2016)Zhang, Zhang, Li, and Qiao}]{Zhang2016Joint}
Zhang K, Zhang Z, Li Z, Qiao Y (2016) Joint face detection and alignment using
  multitask cascaded convolutional networks. IEEE Signal Processing Letters
  23(10):1499--1503

\bibitem[{Zhang et~al.(2012)Zhang, Yan, Liu, Lei, Yi, and Li}]{zhang2012face}
Zhang Z, Yan J, Liu S, Lei Z, Yi D, Li SZ (2012) A face antispoofing database
  with diverse attacks. In: International Conference on Biometrics, pp 26--31

\bibitem[{Zheng et~al.(2024)Zheng, Li, Wu, Wan, Mu, Liu, Ding, and
  Wang}]{zheng2024mfae}
Zheng T, Li B, Wu S, Wan B, Mu G, Liu S, Ding S, Wang J (2024) Mfae: Masked
  frequency autoencoders for domain generalization face anti-spoofing. IEEE
  Transactions on Information Forensics and Security

\bibitem[{Zhou et~al.(2022{\natexlab{a}})Zhou, Yang, Loy, and
  Liu}]{zhou2022conditional}
Zhou K, Yang J, Loy CC, Liu Z (2022{\natexlab{a}}) Conditional prompt learning
  for vision-language models. In: IEEE Conference on Computer Vision and
  Pattern Recognition, pp 16795--16804

\bibitem[{Zhou et~al.(2022{\natexlab{b}})Zhou, Yang, Loy, and Liu}]{Zhou_2022}
Zhou K, Yang J, Loy CC, Liu Z (2022{\natexlab{b}}) Learning to prompt for
  vision-language models. International Journal of Computer Vision
  130(9):2337--2348

\bibitem[{Zhou et~al.(2022{\natexlab{c}})Zhou, Zhang, Yao, Yi, Ding, and
  Ma}]{zhou2022adaptive}
Zhou Q, Zhang KY, Yao T, Yi R, Ding S, Ma L (2022{\natexlab{c}}) Adaptive
  mixture of experts learning for generalizable face anti-spoofing. In: ACM
  International Conference on Multimedia, pp 6009--6018

\bibitem[{Zhou et~al.(2024)Zhou, Zhang, Yao, Lu, Ding, and Ma}]{zhou2024test}
Zhou Q, Zhang KY, Yao T, Lu X, Ding S, Ma L (2024) Test-time domain
  generalization for face anti-spoofing. In: IEEE Conference on Computer Vision
  and Pattern Recognition, pp 175--187

\end{thebibliography}
}

\newpage
\appendix
\section*{Appendix}
\begin{table}[!htbp]
	\footnotesize
	\caption{All the prior descriptions. }
	\label{tabDes}
	\centering 
	\begin{tabular}{p{1cm}<{\centering}p{7cm}
		}
		\hline
		Acronym&Description \\
        \hline
		Des1& a human face with paper surface texture\\
		Des2& a human face with plastic surface texture\\
		Des3& a human face with screens surface texture\\
		Des4& a human face with glossiness or lack of skin-like reflectance properties\\
		Des5& a human face with lack of natural facial movements, such as blinking or subtle micro-expression\\
		Des6& a human face with misalignment or unnatural motion when the spoof medium is manipulated (e.g., hand-held photos or masks)\\
		Des7& a human face with absence of natural depth information, as seen in flat surfaces like printed photos or screens\\
		Des8& a human face with distorted or unnatural depth in 3D masks or molded faces\\
		Des9& a human face with abnormal color distribution, such as oversaturation or uneven illumination\\
		Des10& a human face with differences in skin tone and shading compared to live faces under similar conditions\\
		Des11& a human face with reflection and shadow inconsistencies\\
		Des12& a human face with unnatural reflections caused by glossy materials like screens or masks\\
		Des13& a human face with shadows that do not align with expected lighting conditions\\
		Des14& a human face with visible edges or seams around the spoofing medium, such as cutouts or mask borders\\
		Des15& a human face with blurred or jagged transitions at boundaries, especially in digital forgeries\\
		Des16& a human face with low-quality reproduction\\
		Des17& a human face with pixelation, moiré patterns, or resolution mismatches in screen-based spoofs\\
		Des18& a human face with artifacts from printing or photo degradation in paper-based attacks\\
		Des19& a human face printed on paper, leading to loss of depth and texture fidelity\\
		Des20& a human face with screens or displays, resulting in moiré patterns, pixelation, or unnatural luminance\\
		Des21& a human face with 3D Masks: real faces are replicated using materials like silicone or plastic, which may introduce unnatural textures or geometric distortions\\
		
		\hline
	\end{tabular}
\end{table}

\end{document}